\documentclass[11pt, a4paper]{article}
\usepackage{amsmath, amssymb, amsfonts}
\usepackage{geometry}
\usepackage{graphicx}
\usepackage{booktabs}
\usepackage{array}
\usepackage{microtype}
\usepackage{algorithm}
\usepackage{algpseudocode}
\usepackage{subcaption}
\usepackage{placeins}
\usepackage[colorlinks=true, linkcolor=blue, citecolor=blue, urlcolor=blue]{hyperref}
\hypersetup{pdftitle={Reinforcement Learning for Execution under Dynamic Fees in a Closed-Loop DEX Simulator}, pdfauthor={Wen-Ting Wang}}

\newcommand{\bps}{\,\mathrm{bps}}

\title{\textbf{Reinforcement Learning for Execution under Dynamic Fees\\
in a Closed-Loop DEX Simulator}}
\author{Wen-Ting Wang}
\date{July 2026\\
{\small Email: \texttt{egpivo@gmail.com}}}

\begin{document}

\maketitle

\begin{abstract}
Trader-facing dynamic fees are increasingly proposed for automated market makers (AMMs),
but historical data do not identify how order flow would respond: trader-facing fees do
not vary, trader types are latent, and a replayed tape is not a sequential decision
environment. We therefore construct a minimal closed-loop simulator in which the missing
signal exists by construction: two constant-product pools repriced by an
equilibrium-inspired dynamic-fee rule, fee-sensitive noise flow, and closed-form
CEX--AMM arbitrage. Equilibrium is used as a closure principle, not as an object the
trader learns. Against a tuned benchmark ladder of schedule, planning, lookahead, and
tabular policies, a small DQN is the only evaluated valid policy whose paired improvement
over tuned one-step routing excludes zero. On a reserved final block of 1{,}000 seeds
with completion forced to 1.0 for every policy, it reduces implementation shortfall under
every tested intra-step ordering, by $13.3\bps$ of order notional under the pre-specified
agent-last ordering, and the edge is concentrated in, and learned from, dynamic-fee
environments: under constant fees the paired difference is indistinguishable from zero.
The result is model-conditioned counterfactual evidence about execution control in AMMs,
not evidence about historical traders, equilibrium play, or deployable profit.
\end{abstract}

\medskip
\noindent\textbf{Keywords:} automated market makers; decentralized exchanges; dynamic fees;
optimal execution; reinforcement learning; market simulation.

\section{Introduction}
\label{sec:intro}

In decentralized exchange research, it is of interest to know whether trader-facing dynamic
fees protect liquidity providers, and, dually, whether execution agents can defend themselves
against fee rules that reprice as they trade. A commonly used approach is empirical: estimate
responses from historical swaps and liquidity events. However, historical identification stops
at specific margins. In our companion event study of the Uniswap protocol-fee
switch~\cite{companion}, the
LP-side response to take-rate changes is identified by the design; the trader-facing question
is not, because trader-facing fees do not vary, trader types are latent, and the routing
decision set a trader faced at each block is unobserved. Questions of the form ``what would
order flow do if fees moved against it'' are counterfactuals the record cannot answer.

A second approach applies reinforcement learning to trade
execution~\cite{nevmyvaka2006,hendricks2014,ning2021}, in our case by training an agent
on the historical tape. In
preliminary experiments we ran that program across four data regimes and closed it with a
negative result: the
historical tape does not identify an action-dependent transition kernel for counterfactual
policies. A replayed agent can condition on history, but its actions cannot change
subsequent states, and each apparent pocket of adaptive
headroom was traced to a measurable artifact by pre-specified audits. The binding constraint
was not sample size but the information content of the signal.

To address this issue, we stop asking the tape for a signal it does not contain and instead
construct a minimal market in which the signal exists by construction: a closed-loop
simulator in which an execution trader's actions move pool inventory, which moves quotes, which
moves a dynamic fee rule, noise-flow routing, and an arbitrageur's response. Two design
commitments separate this from a generic market game. First, the fee rule is
\emph{equilibrium-inspired closure}: a linearization of the approximate Nash fee structure
derived for competing constant-function market makers by Baggiani, Herdegen, and
S\'anchez-Betancourt~\cite{baggiani2026competition,baggiani2025optimal}, whose role is to make
the environment defend itself, not to certify equilibrium play. Second, evaluation discipline is
inherited from that audit protocol: baseline hyperparameters are validation-selected on
200 seeds and the DQN checkpoint on 50, every claim is
gated behind artifact diagnostics, and the headline number comes from a seed block untouched by
any development decision.

The research question is narrow: under an equilibrium-inspired dynamic-fee closure, can a
model-free learner reduce implementation shortfall relative to strong tuned
execution heuristics, and does the advantage survive when the market reprices, reroutes, and
defends itself? The answer, summarized in Figure~\ref{fig:final} (with the full
agent-first benchmark ladder in Appendix~\ref{app:ladder}), is yes under explicit
conditions: a small DQN beats a
validation-tuned one-step lookahead benchmark, frozen before the priority experiments, by
$5.6$ to $14.9\bps$ of order notional depending on
intra-step priority ordering, at completion forced to $1.0$, on the reserved final block, while
deterministic and stochastic shallow planners fail to close the gap and the best tabular
learner's paired difference from lookahead has a confidence interval containing zero. The
advantage is concentrated in dynamic-fee environments; under constant fees the paired
difference between learner and heuristic is not distinguishable from zero.

A closely related line develops tractable stochastic-control solutions for execution and
speculation on constant-product markets: Cartea, Drissi, and
Monga~\cite{cartea2023execution} solve for optimal strategies when the market's response
to the trader is the pool's price impact. The present setting asks a complementary
question: when fees reprice, flow reroutes, and an arbitrageur responds to the trader's
footprint, comparably tractable solutions are not available, and the contribution is an
evaluation discipline for model-free control in that closed loop rather than a solvable
model.

Section~\ref{sec:method} states the evaluation method independently of the market model;
Section~\ref{sec:study} instantiates it on a dynamic-fee DEX and reports the full simulation
study; Section~\ref{sec:scope} delimits what the result does and does not support. The
implementation (the simulator, all policies, and the training and evaluation pipeline) is
available at \url{https://github.com/egpivo/amm-lab}.

\section{Closed-Loop Execution Method}
\label{sec:method}

The reusable method is a discipline, not a learner: construct an action-responsive
environment with declared transition semantics; impose common completion semantics on every
policy; compare against a tuned policy ladder rather than a single baseline; freeze all
development choices before touching a reserved evaluation block; and, when the environment
is perturbed, distinguish transfer failure of a frozen policy from failure after re-solving.
This section states each component for a generic market simulator; Section~\ref{sec:study}
instantiates every symbol. The method is reusable as a simulator-evaluation protocol;
neither the learned policies nor the empirical gains are claimed to transfer across market
specifications, and the results of Section~\ref{sec:study} are specific to the
instantiated market and assumptions.

\subsection{Action-dependent market transition}
\label{sec:transition}

The environment is a finite-horizon MDP $(\mathcal{S}, \mathcal{A}, T, r, H)$ whose
transition composes the controller's action with a sequence of market-response operators in
a fixed, declared order:
\begin{equation}
\label{eq:transition-generic}
s_{t+1} = \big(\Phi^{(m)}_{\mathrm{mkt}} \circ \cdots \circ \Phi^{(1)}_{\mathrm{mkt}}
\circ \Phi_{\mathrm{agent}}(a_t)\big)(s_t;\, \xi_t),
\end{equation}
where $\xi_t$ collects the step's exogenous shocks. Eq.~\eqref{eq:transition-generic}
shows the agent-first composition; other priority regimes permute the same operators. Two semantic commitments matter more
than the specific operators. First, the policy chooses $a_t$ from the observation of $s_t$;
the controller leg is then priced by the market state at its declared position in the
operator sequence. If market operators act first, the chosen action executes against the
resulting intermediate state, which was not observed when $a_t$ was selected. No policy
observes outputs produced later than its decision, so future-information leakage is
excluded by construction rather than by convention. Second, the controller's intra-step position is an
explicit design variable of $T$, not an accident of implementation: the controller may act
before the market operators, after them (on stale quotes), or at a randomized position;
Section~\ref{sec:artifacts} shows it is quantitatively load-bearing.

\subsection{Execution objective and strict completion}
\label{sec:objective}

The task is a target order of $Q$ units over $H$ steps; $R_t$ denotes remaining inventory
($R_0 = Q$) and $S_0$ the arrival price. With venue cost functions $c_i(\cdot)$ (venue cash cost of an immediate trade, including
venue fees but excluding the separately charged transaction cost $g$; instantiated in
Section~\ref{sec:market}), per-trade transaction cost $g$, and $q_{i,t}$ the quantity sent to venue $i$, the per-step reward is the negative
normalized execution cost
\begin{equation}
\label{eq:reward}
r_t = -\frac{1}{QS_0}\Big[\sum_{i} \big(c_i(q_{i,t}) - q_{i,t} S_0\big)
 + g\, n_t\Big] \;-\; \frac{C^{\mathrm{term}}_H}{QS_0}\,\mathbf{1}\{t = H-1\},
\end{equation}
where $n_t$ is the number of venues touched and the terminal cost is defined by the
completion rule:
\begin{equation}
\label{eq:cterm}
C^{\mathrm{term}}_H =
\begin{cases}
\rho\, S_0 R_H, & \text{standard rule}, \\[2pt]
C^{\mathrm{force}}(R_H) - S_0 R_H + g\, n^{\mathrm{force}}_H, & \text{forced-terminal rule},
\end{cases}
\end{equation}
with $\rho$ a penalty rate, $R_H$ the inventory left after the final chosen action, and
$C^{\mathrm{force}}(R_H)$ the execution cash cost of the forced residual leg: after the
final chosen controller action executes at its configured intra-step position, any
residual inventory is liquidated immediately against the same post-action venue state and
fee snapshot, before any subsequent operators of that step, including venue fees but
excluding the separately charged transaction cost; the forced leg incurs its own
per-venue charge $g$ on the $n^{\mathrm{force}}_H$ venues it touches, in addition to any
charge on the chosen action. Both rules are applied identically to every
policy. For an episode with exogenous shock path $\xi = \xi_{0:H-1}$ evaluated under
completion rule $c$, the episode score is
$L_c(\pi; \xi) = -10^{4} \sum_{t=0}^{H-1} r_t(\pi; \xi)$ in bps of notional. Under the
forced-terminal rule every unit is actually executed, so $L_{\mathrm{force}}$ is the
realized implementation shortfall~\cite{perold1988}, written $\mathrm{IS}(\pi; \xi)$;
under the standard rule $L_{\mathrm{std}}$ is a \emph{penalized execution score}, not
literal shortfall, because residual inventory is charged a synthetic terminal penalty
rather than executed. All headline comparisons use the forced-terminal rule. With mean
$\mu_\pi = \mathbb{E}_\xi[L_c(\pi; \xi)]$ under the rule in force (for deterministic
evaluation policies the expectation is over exogenous shocks only; for randomized
policies it additionally integrates over the policy's seeded internal randomization), a
policy maximizes
$J(\pi) = \mathbb{E}_\pi\big[\sum_{t=0}^{H-1} r_t\big] = -10^{-4} \mu_\pi$
(undiscounted; the horizon is in the state). The comparison estimand under ordering $o$ is
$\Delta_o = \mathbb{E}_\xi\big[\mathrm{IS}(\pi_{\mathrm{DQN}}, o; \xi) -
\mathrm{IS}(\pi_{\mathrm{LA}}, o; \xi)\big]$, estimated by the seed average
$\hat{\Delta}_o = n^{-1} \sum_{j=1}^{n} \big[\mathrm{IS}_{j,\mathrm{DQN},o} -
\mathrm{IS}_{j,\mathrm{LA},o}\big]$. Under the forced-terminal rule
completion is $1.0$ by construction and no policy can trade shortfall against unfilled
inventory.

Because $\gamma = 1$ and Eq.~\eqref{eq:reward} contains no shaping terms, the episode
return sums to the episode's total normalized cost:
\begin{equation}
\label{eq:telescope}
\sum_{t=0}^{H-1} r_t = -\frac{1}{QS_0}\Big[\underbrace{\textstyle\sum_t \sum_i
\big(c_i(q_{i,t}) - q_{i,t}S_0\big)}_{\text{execution premium}}
 + \underbrace{\textstyle\sum_t g\,n_t}_{\text{transaction costs}}
 + \underbrace{C^{\mathrm{term}}_H}_{\text{terminal}}\Big],
\end{equation}
with $C^{\mathrm{term}}_H$ as in Eq.~\eqref{eq:cterm}; taking expectations,
$J(\pi) = -10^{-4}\,\mu_\pi$, and the identity holds under both completion rules. Within a fixed completion rule, maximizing $J(\pi)$ is therefore identical
to minimizing the corresponding episode-score accounting identity, terminal cost included, with no
discount distortion in which early costs outweigh late ones; under-completion is not
omitted from the objective, and the completion audit tests whether the specified penalty is
sufficient. A training--evaluation caveat follows from the same identity: a completion-rule switch
between training and evaluation must be disclosed as an out-of-training-semantics
evaluation and applied uniformly to every policy, and its effect on the learned policy must
then be audited directly (the completion audit of Section~\ref{sec:artifacts}).

\subsection{Observation and policy classes}
\label{sec:obsclass}

The simulator exposes a raw decision-time observation $\tilde{o}_t$ containing the
current-state variables a market participant could see; policies differ in how much of it
they consume. Heuristic baselines and planners use $\tilde{o}_t$ directly. Learners act on
compressions: a feature map $o_t = \phi(\tilde{o}_t) \in \mathbb{R}^{d}$ for function
approximation, and a coarser discretization $\psi: \mathbb{R}^{d} \to \{1, \dots, M\}$
for tabular control. No policy receives
future information (a schema test whitelists the observation components), and the
information asymmetry runs \emph{against} the learner, since the heuristics see strictly
more of the current state. The action-value of the optimal policy,
\begin{equation}
\label{eq:qstar}
Q^\star_t(s, a) = \mathbb{E}\Big[r_t + \max_{a'} Q^\star_{t+1}(s_{t+1},\, a')
\,\Big|\, s_t = s,\ a_t = a\Big], \qquad Q^\star_H \equiv 0,
\end{equation}
with $s_{t+1} = T(s, a;\, \xi_t)$ and $r_t$ the reward generated by taking $a$ in $s$,
defined on the full state, is (the negative of) the expected \emph{cost-to-go} of the
remaining execution problem, the dynamic-programming view of execution costs introduced
by Bertsimas and Lo~\cite{bertsimas1998}; every early decision is priced against this long-term
quantity, which is how the terminal cost and late-horizon dynamics reach step-0 behavior.
Learners are observation-based and do not approximate the fully observed $Q^\star$
directly; with $\gamma = 1$, a finite horizon, and time in the state, they target the same
undiscounted finite-horizon cost-to-go objective within their observation class. The
compressed observation is not assumed to be Markov-sufficient: Eq.~\eqref{eq:qstar}
defines optimality on the full simulator state, and the DQN applies a Bellman-style
approximation on $o_t$ without a convergence or optimality claim for the compressed
process.

\subsection{Benchmark ladder}
\label{sec:ladderdef}

A single baseline invites overstatement; the method compares every learner against a ladder
whose rungs each rule out one alternative explanation of an apparent edge:

\emph{Schedule baselines} (fastest-feasible liquidation; TWAP on the better quote; a
myopic best-quote router; a random policy; a fee-aware TWAP choosing among same-pace
actions by estimated all-in cost) rule out value that requires no adaptivity. Every rung
of the ladder, schedule baselines included, acts in the controller's action set
$\mathcal{A}$; schedule baselines are fixed rules within it, with their pacing projected
onto the available actions (Section~\ref{sec:impl}). Writing $S_t$ for the
current reference price, $q_i(a) = \varphi_i(a)\, R_t$ for the quantity action $a$ sends
to venue $i$ (with $\varphi_i(a)$ the corresponding fraction of remaining inventory), and
$n(a)$ for the number of venues it touches, \emph{tuned one-step control} is the benchmark
every learner must beat:
\begin{equation}
\label{eq:lookahead}
\pi_{\mathrm{LA}}(\tilde{o}_t) = \arg\min_{a \in \mathcal{A}}\;
\underbrace{\sum_i \big(\hat{c}_i(q_i(a)) - q_i(a)\, S_t\big) + g\, n(a)}_{\text{immediate
premium}}
\;+\; U_t(a),
\end{equation}
where $\hat{c}_i$ is the exact quote curve rebuilt from $\tilde{o}_t$ and the urgency
term evaluates a single urgency function at the \emph{post-action} inventory,
$U_t(a) = C_{\kappa_{\mathrm{LA}}}\big(H - t - 1,\, S_t,\,
R_t(1 - \sum_i \varphi_i(a))\big)$, with
\begin{equation}
\label{eq:carry}
C_{\kappa}(\ell, S, R) =
\begin{cases}
\kappa\, \rho\, S\, R \,/\, \ell, & \ell \ge 1, \\[4pt]
\rho\, S\, R, & \ell = 0,
\end{cases}
\end{equation}
where $\ell$ counts the decision opportunities remaining \emph{after} the one being
priced; the $\ell = 0$ branch charges the full mark-to-market penalty without $\kappa$
(a planner-side terminal proxy, not the environment's forced-terminal cost).
$C_{\kappa}$ is a mark-to-market proxy, an immediate-cost-versus-deferral trade-off in
the spirit of Almgren and Chriss~\cite{almgren2001}, calibrated to the terminal-penalty
scale: it prices deferred inventory at the current reference price, not the arrival
price $S_0$ used by the terminal cost itself. The coefficient is selected per policy
class ($\kappa_{\mathrm{LA}}$ here; per-depth values for the deterministic planners;
$\kappa_{\mathrm{roll}}$ at the rollout leaf), each validation-tuned per market mode
under the default ordering and frozen before cross-ordering evaluation, with the grid
extended whenever the optimum sits on its edge. \emph{Multi-step planners} rule out value
reachable by model-based search: a deterministic certainty-equivalent $K$-step planner on
a forward model $\hat{T}$ rebuilt from $\tilde{o}_t$ at every decision (mean shocks,
martingale reference price; the chance nodes are collapsed to their means, so this is
certainty-equivalent planning rather than expectimax; at its deepest ply it applies the
same post-action carry $C_{\kappa_K}$, and a model state reaching the horizon inside the
tree is charged the full proxy), and a stochastic rollout planner
in the sense of~\cite[\S 8.10]{sutton2018} (decision-time Monte Carlo value estimates of
each current action under a fixed rollout policy) that samples the shocks instead,
\begin{equation}
\label{eq:rollout}
\hat{V}(a) = \mathrm{cost}(a) + \frac{1}{N}\sum_{n=1}^{N}
\Big[\sum_{k=1}^{K_t-1} \mathrm{cost}\big(\pi_{\mathrm{LA}}(\hat{\tilde{o}}^{(n)}_{t+k})\big)
\;+\; C_{\kappa_{\mathrm{roll}}}\big(\ell_{t},\, \hat{S}^{(n)}_{t+K_t},\,
\hat{R}^{(n)}_{t+K_t}\big)\Big],
\end{equation}
over $N$ rollouts of $\hat{T}$ with shocks drawn from the simulator's own distributions but
from RNG streams keyed to the decision index, never to the episode seed, so no
planner sees realized future shocks. Here $K_t = \min\{K,\, H - t\}$ is the effective
depth, so near the horizon rollouts truncate at $H$ rather than beyond it (they also stop
early if model inventory is exhausted, where the leaf value vanishes with
$\hat{R} = 0$); the leaf argument $\ell_t = \max\{H - t - K_t - 1,\, 0\}$ applies the
same post-decision convention as $U_t$, treating the leaf decision as the one being
priced; $\mathrm{cost}(\cdot)$ is the immediate-premium term
of Eq.~\eqref{eq:lookahead} evaluated on the model state, and
$\hat{\tilde{o}}^{(n)}_{t+k}$, $(\hat{S}^{(n)}_{t+K_t}, \hat{R}^{(n)}_{t+K_t})$ denote the
model observation and the leaf price and inventory along rollout $n$. The urgency function
$C_{\kappa}$ of Eq.~\eqref{eq:carry} thus serves the lookahead and planner rungs through
one functional form: the lookahead ranks actions with it, the deterministic planners
apply it at their deepest ply, and the truncated rollout uses it once as the leaf
continuation value, evaluated at the full leaf inventory (the leaf decision is implicitly
a wait), each with its own tuned coefficient rather than a shared parameter. Because the accumulated rollout
cost sums only $\mathrm{cost}(\cdot)$ and excludes $C_{\kappa}$ ($U$ enters the rollouts
only through the continuation policy's action choices), adding it once at the leaf does
not double-count. \emph{Learners}: tabular on-policy Monte Carlo control
with $\varepsilon$-greedy exploration (a constant-$\alpha$, every-visit variant of the
on-policy method of~\cite[\S 5.4]{sutton2018}, with no convergence claim carried over),
\begin{equation}
\label{eq:mc}
\hat{Q}\big(\psi(o_t), a_t\big) \leftarrow \hat{Q}\big(\psi(o_t), a_t\big)
 + \alpha\,\big(G_t - \hat{Q}(\psi(o_t), a_t)\big),
\qquad G_t = \textstyle\sum_{u \ge t} r_u,
\end{equation}
with step size $\alpha$ and return-to-go $G_t$, and a DQN following Mnih
et al.~\cite{mnih2015}, as applied to optimal execution in~\cite{ning2021}, trained on
the Huber loss $\ell_\delta$ against the standard
target
\begin{equation}
\label{eq:dqn}
y_t = r_t + (1 - d_t)\, \max_{a'} Q_{\theta^-}(o_{t+1}, a'), \qquad
\mathcal{L}(\theta) = \mathbb{E}_{(o_t, a_t, r_t, o_{t+1}, d_t) \sim \mathcal{D}}\;
\ell_\delta\big(Q_\theta(o_t, a_t) - y_t\big),
\end{equation}
with replay buffer $\mathcal{D}$, target network $\theta^-$, and $d_t$ the
episode-termination indicator. Finally, a
\emph{clairvoyant} coordinate descent over full action sequences
$\mathbf{a} \in \mathcal{A}^H$ evaluated on the true episode, which does see future
shocks and is not a valid policy, provides an achieved hindsight reference
$\mathrm{IS}(\hat{\mathbf{a}}_{\mathrm{CD}};\, \xi_{0:H-1})$, where
$\hat{\mathbf{a}}_{\mathrm{CD}}$ is the sequence found by the search, without certifying
an optimal clairvoyant bound.

\subsection{Training, freeze, and evaluation protocol}
\label{sec:protocol}

Algorithm~\ref{alg:pipeline} fixes the pipeline for one environment configuration. Its
guarantees are procedural: checkpoint selection touches only a validation block; baselines
are tuned on a superset of that block; every design choice is frozen and content-hashed
before the reserved final block is evaluated; and all comparisons are paired per seed
against the tuned one-step benchmark. Because learner design typically iterates while
development seeds are visible (a general evaluation hazard in deep
RL~\cite{henderson2018}), the reserved block $\mathcal{S}_{\mathrm{final}}$ (never
used in any training, tuning, or development run) is the only source of headline
numbers. Confidence intervals resample the $n$ seed-level policy pairs with replacement
(percentile bootstrap; $B = 4{,}000$ replications for the headline comparison of
Figure~\ref{fig:final}, $2{,}000$ for other figure whiskers) and are pointwise $95\%$ intervals; no multiplicity correction is applied
across secondary orderings, ladder rungs, or the robustness battery, which is acceptable
only because the agent-last cell is the pre-specified primary comparison. All intervals
condition on the frozen selected checkpoint and its recorded training seed: they quantify
variation across market shock paths, not variation induced by retraining the learner.

\begin{algorithm}[!htbp]
\caption{Training, selection, freeze, and evaluation (one market mode; the learner is
trained per declared ordering, baselines tuned once under the mode's default ordering)}
\label{alg:pipeline}
\begin{algorithmic}[1]
\Require config (mode $\times$ ordering); training and evaluation completion rules
$(c_{\mathrm{train}}, c_{\mathrm{eval}})$, disclosed when they differ
(Section~\ref{sec:objective}); seed blocks with
$\mathcal{S}_{\mathrm{train}}, \mathcal{S}^{B}_{\mathrm{val}},
\mathcal{S}_{\mathrm{final}}$ pairwise disjoint and
$\mathcal{S}^{L}_{\mathrm{val}} \subseteq \mathcal{S}^{B}_{\mathrm{val}}$;
re-solving set $\mathcal{R}_{\mathrm{resolve}}$
\State \textbf{Train:} for $e = 1, \dots, E$: rollout $\varepsilon$-greedy $Q_\theta$
under $c_{\mathrm{train}}$ on seed $s_e \in \mathcal{S}_{\mathrm{train}}$; update by
Eq.~\eqref{eq:dqn}; $\theta^- \leftarrow \theta$ every $\tau$ steps
\State \textbf{Select:} every $E_{\mathrm{eval}}$ episodes: checkpoint
$\leftarrow \arg\min$ mean score under $c_{\mathrm{train}}$ on $\mathcal{S}^{L}_{\mathrm{val}}$
\State \textbf{Tune lookahead:} $\kappa \leftarrow \arg\min$ mean score on
$\mathcal{S}^{B}_{\mathrm{val}}$ under the mode's default ordering and the declared
tuning completion rule
\State \textbf{Fix planner rungs:} depths $K \in \{2, 3\}$ pre-declared; carry re-tuned
per depth on $\mathcal{S}^{B}_{\mathrm{val}}$
\State \textbf{Tune stochastic planner:} $(K, N, \kappa_{\mathrm{roll}}) \leftarrow
\arg\min$ mean score on $\mathcal{S}^{B}_{\mathrm{val}}$
\State \textbf{Audit evaluation semantics:} re-evaluate the validation grid under
$c_{\mathrm{eval}}$; record whether the selected $\kappa$ changes
\State \textbf{Freeze priority comparator:} freeze the selected baseline parameters
before any cross-ordering or final-block evaluation
\State \textbf{Freeze specification:} serialize and hash the checkpoint, completion
rules $(c_{\mathrm{train}}, c_{\mathrm{eval}})$, ordering, baseline parameters, seed
roles, software versions, and commands
\State \textbf{Evaluate:} all frozen policies under the common rule $c_{\mathrm{eval}}$
on $\mathcal{S}_{\mathrm{final}}$; report mean score, paired CIs vs.\ $\pi_{\mathrm{LA}}$
\State \textbf{Finalize manifest:} append hashes of the result files to the frozen
specification
\State \textbf{Diagnose:} $\forall\, M'$: frozen transfer; if
$M' \in \mathcal{R}_{\mathrm{resolve}}$: re-solve
\end{algorithmic}
\end{algorithm}

\subsection{Transfer versus re-solving}
\label{sec:transfer}

$Q^\star$ in Eq.~\eqref{eq:qstar} is defined relative to \emph{both} the transition and
reward semantics of $\mathcal{M} = (\mathcal{S}, \mathcal{A}, T, r, H)$: the cost-to-go
prices the specific market response and cost accounting that follow each action. Every
robustness perturbation is a change $\mathcal{M} \to \mathcal{M}'$ (some perturb $T$
alone, such as priority ordering; others perturb $r$ or the terminal semantics, such as gas
and the completion rule), and a policy greedy with respect to $Q^\star_{\mathcal{M}}$ has no
optimality guarantee under $\mathcal{M}'$. The protocol therefore distinguishes two
questions that a single robustness number conflates: whether the \emph{frozen} policy's
advantage survives evaluation under $\mathcal{M}'$ (transfer), and whether the advantage
exists at all under $\mathcal{M}'$ when the learner is retrained there (re-solving). The classification is fixed in advance rather
than after seeing results: frozen transfer is reported for every perturbation; the
pre-specified re-solving set $\mathcal{R}_{\mathrm{resolve}}$ contains fee mode and
intra-step priority ordering; scalar nuisance perturbations (gas, arbitrage speed, noise
scale, fee coefficients) and additive appendix stress layers are transfer-only unless
otherwise stated. A transfer failure
paired with a re-solving success is consistent with policy--environment mismatch rather
than an absence of exploitable structure under $\mathcal{M}'$; Section~\ref{sec:artifacts} shows
exactly this pattern for intra-step priority.

\FloatBarrier
\section{Simulation Study: Dynamic-Fee DEX}
\label{sec:study}

\subsection{Market instantiation}
\label{sec:market}

\begin{figure}[!htbp]
\centering
\includegraphics[width=0.72\textwidth]{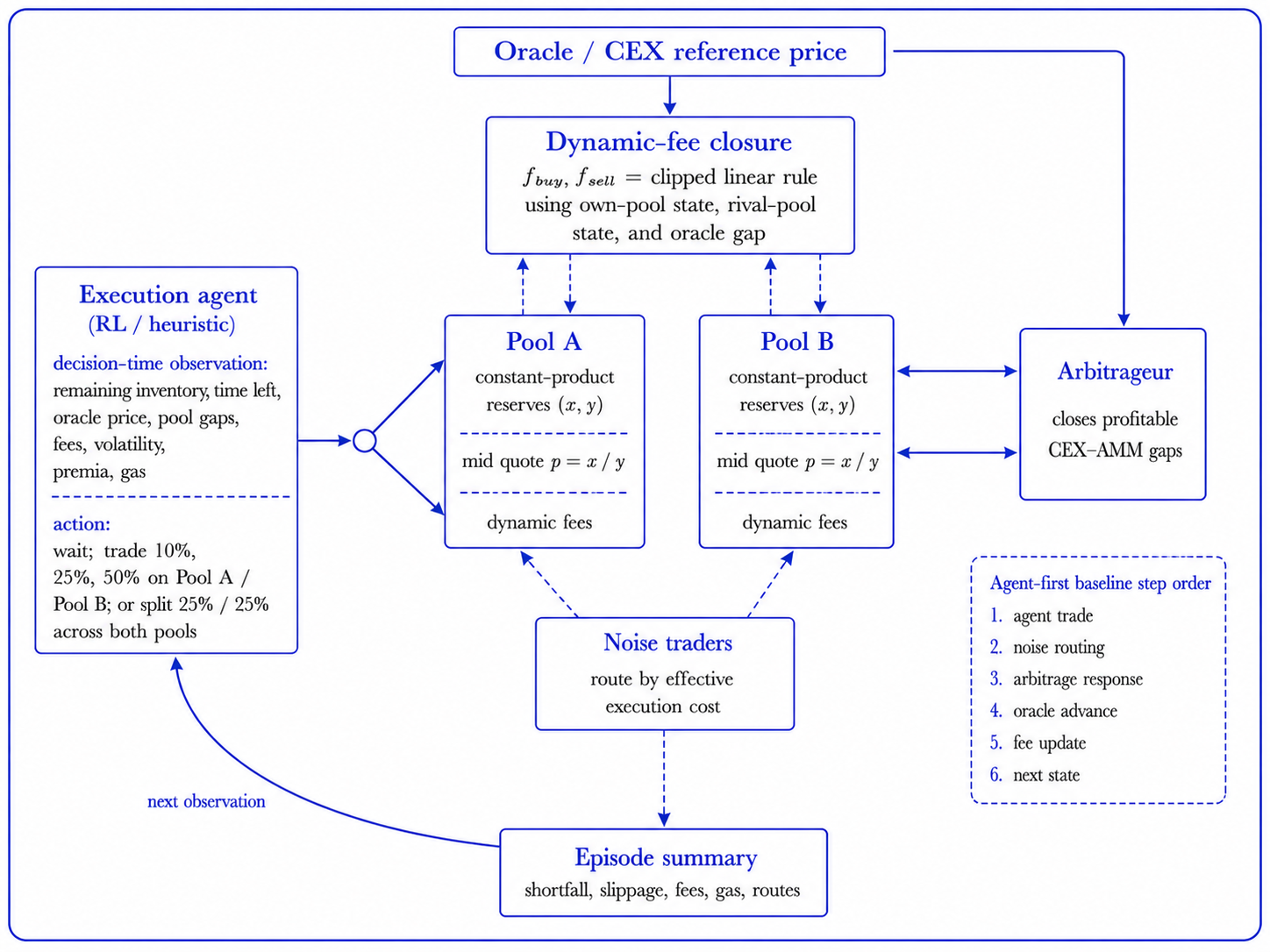}
\caption{The closed-loop market. The execution agent trades a target order across two
constant-product pools; the dynamic-fee closure reprices both pools from own-pool state,
rival-pool state, and the oracle gap; noise traders route by effective cost; an
arbitrageur closes profitable CEX--AMM gaps. The agent box lists the decision-time
observation $\tilde{o}_t$ (Section~\ref{sec:obsclass}); the step order shown is the
agent-first baseline of Eq.~\eqref{eq:transition}, and the agent-last and randomized
configurations permute the agent leg (Section~\ref{sec:transition}).}
\label{fig:market}
\end{figure}

Figure~\ref{fig:market} sketches the instantiation. One risky asset $Y$ trades against a
numeraire $X$ on two stylized constant-product pools in the Uniswap
lineage~\cite{angeris2019uniswap} and an external venue. The external (CEX) mid
price $S_t$ follows a seeded geometric Brownian motion with zero drift. Each pool $i$ holds
reserves $(x_i, y_i)$ and charges directional fees
$(f^{\mathrm{buy}}_i, f^{\mathrm{sell}}_i)$. Fees accrue outside reserves, so the invariant
$x_i y_i$ is preserved exactly by every trade; this matches the fee treatment
in~\cite{baggiani2026competition} and makes conservation unit-testable.

Pool $i$ updates fees each step by a clipped linear rule,
\begin{equation}
\label{eq:fee}
f^{\mathrm{buy}}_i = \Big[\bar{f} + a_{\mathrm{own}} z_i + a_{\mathrm{riv}} z_{-i}
 + a_{\mathrm{orc}} m_i\Big]_{f_{\min}}^{f_{\max}}, \qquad
f^{\mathrm{sell}}_i = \Big[\bar{f} - a_{\mathrm{own}} z_i - a_{\mathrm{riv}} z_{-i}
 - a_{\mathrm{orc}} m_i\Big]_{f_{\min}}^{f_{\max}},
\end{equation}
where, with mid quote $p_{i,t} = x_{i,t}/y_{i,t}$,
\[
z_{i,t} = \frac{x_{i,t} - y_{i,t} S_t}{x_{i,t} + y_{i,t} S_t} \in [-1, 1],
\qquad
m_{i,t} = \frac{p_{i,t} - S_t}{S_t},
\]
are the oracle-valued inventory imbalance and the mid--oracle relative gap (so a pool
priced below the oracle has $m_i < 0$ and, with $a_{\mathrm{orc}} > 0$, a lower buy fee),
$\bar{f}$ is the base fee, $(a_{\mathrm{own}}, a_{\mathrm{riv}}, a_{\mathrm{orc}})$ are
the response coefficients, and $[\cdot]_{f_{\min}}^{f_{\max}}$ clips to the fee bounds
(values in Table~\ref{tab:provenance}). All fee quantities and coefficients are decimal
fractions; $30\bps = 0.003$. Because $p_{i,t} = x_{i,t}/y_{i,t}$, the two own-pool inputs
are mechanically linked, $z_{i,t} = m_{i,t}/(2 + m_{i,t})$, so they are not separately
identified in this minimal CPMM: both terms move the fee through a single own-pool channel
(alongside the rival term), and $a_{\mathrm{own}}$ and $a_{\mathrm{orc}}$ are retained to
mirror the cited parameterization, not interpreted as independent structural effects. The
functional form follows the two findings
of~\cite{baggiani2025optimal,baggiani2026competition}: the cited optimal-fee solutions are
approximately linear in inventory and track external price changes, and under competition
their switching boundary depends on both the oracle and the rival's quote. We do not solve
the coupled PDE system; the rule's only job is closure, and whether it behaves in the
expected two-regime way is itself a diagnostic (Section~\ref{sec:artifacts}). Fee design
at DEXs is a wider literature than the equilibrium line used here (fee increases can
raise volume through equilibrium liquidity~\cite{hasbrouck2022fees}; pricing-function
design has its own optimization program~\cite{bergault2024amm}); Eq.~\eqref{eq:fee} is
one representative closure, not a claim about optimal fee design.

Noise traders submit buy and sell volume each step (positive multiplicative noise around
fixed median scales) and route it across pools by a softmax over effective execution cost, so
cheaper pools attract more flow without perfect routing, in line with cost-sensitive but
imperfect DEX order flow~\cite{lehar2024uniswap,capponi2021dex}; the logistic form itself
is a researcher-chosen reduced form (Appendix~\ref{app:repro}), and the citations
motivate cost sensitivity, not the functional form. An arbitrageur observes the oracle
and both pools and trades to the fee-adjusted no-arbitrage
ratio~\cite{angeris2020oracles} in closed form whenever
profit exceeds gas, with an optional per-step participation probability standing in for
latency. The CEX--AMM structure follows Campbell, Bergault, Milionis, and
Nutz~\cite{campbell2025optimal}: the external venue is costly, fundamental flow routes by
marginal all-in price, and arbitrageurs close the gap; the cost this rebalancing flow
imposes on the pools is related to the loss-versus-rebalancing channel of Milionis et
al.~\cite{milionis2022lvr}.

The full market state at step $t$ is
\begin{equation}
\label{eq:state}
s_t = \big(t,\; R_t,\; a_{t-1},\; S_{t-w:t},\; (x_{A,t}, y_{A,t}, f^{\mathrm{buy}}_{A,t},
f^{\mathrm{sell}}_{A,t}),\; (x_{B,t}, y_{B,t}, f^{\mathrm{buy}}_{B,t},
f^{\mathrm{sell}}_{B,t})\big),
\end{equation}
where $S_{t-w:t}$ is the trailing oracle-price window ($w$ the volatility window); the
Markov property holds with respect to this state. The action set
$\mathcal{A} = \{a_0, \dots, a_7\}$ maps to per-pool fractions of remaining inventory,
\[
\big(\varphi_A(a), \varphi_B(a)\big) \in \{(0,0),\, (.1,0),\, (0,.1),\, (.25,0),\,
(0,.25),\, (.5,0),\, (0,.5),\, (.25,.25)\},
\]
and the executed quantity on pool $i$ is $q_{i,t} = \varphi_i(a_t)\, R_t$. A buy of $q$ on pool $i$ costs, with fees outside
reserves,
\begin{equation}
\label{eq:cost}
c_i(q) = \frac{1}{1 - f^{\mathrm{buy}}_i}\cdot\frac{x_i\, q}{y_i - q},
\qquad x_i \leftarrow x_i + (1 - f^{\mathrm{buy}}_i)\, c_i(q), \quad y_i \leftarrow y_i - q,
\end{equation}
plus gas $g$ per pool touched; Eq.~\eqref{eq:cost} instantiates the venue cost functions of
Section~\ref{sec:objective}. Sells (used by the noise and arbitrage legs) mirror
Eq.~\eqref{eq:cost} with the fee withheld from the numeraire proceeds, so the invariant is
preserved exactly in both directions; Appendix~\ref{app:repro} gives the closed form. The transition Eq.~\eqref{eq:transition-generic} instantiates
with four market operators,
\begin{equation}
\label{eq:transition}
s_{t+1} = \big(\Phi_{\mathrm{fee}} \circ \Phi_{\mathrm{orc}} \circ \Phi_{\mathrm{arb}}
\circ \Phi_{\mathrm{noise}} \circ \Phi_{\mathrm{agent}}(a_t)\big)(s_t;\, \xi_t),
\end{equation}
where $\xi_t$ collects noise volumes, arbitrage participation, and the GBM increment;
$\Phi_{\mathrm{orc}}$ advances the oracle, and $\Phi_{\mathrm{fee}}$ applies
Eq.~\eqref{eq:fee} \emph{last}, producing the fees that are part of $s_{t+1}$: the
$z_i$, $z_{-i}$, and $m_i$ entering Eq.~\eqref{eq:fee} at that point are computed from the
post-agent, post-noise, post-arbitrage reserves and the advanced oracle $S_{t+1}$, so
$f_{i,t+1}$ is a function of the end-of-step state. The fees
active when $a_t$ executes are therefore fixed before the step begins, as
Section~\ref{sec:transition} requires. The ordering configurations swap
$\Phi_{\mathrm{agent}}$ to after $\Phi_{\mathrm{arb}}$ (agent-last) or randomize the
position with a seeded coin per step.

\subsection{Parameterization and provenance}
\label{sec:params}

The task is a buy of $Q = 50\,Y$ (5\% of one pool's initial risky-asset reserve, 2.5\% of
aggregate risky reserves) over $H = 50$ one-minute steps.
Table~\ref{tab:provenance} (Appendix~\ref{app:repro}) catalogues each parameter group
with its provenance class and points to the full constants where needed. What is anchored to data is limited to scales, chosen to match common on-chain
magnitudes without formal estimation: the base fee, the volatility range, the
order-size-to-depth ratio, and gas-to-notional. The fee coefficients and clips
are design parameters, literature-inspired in form but researcher-chosen in value, not
estimates. Everything else (the GBM oracle, the fee rule, flow elasticities, arbitrage
behavior) is model-conditioned, and every number below is counterfactual output of this
model, not a measurement of history.

Pool depths are fixed
over the horizon in the baseline environment, the assumption
of~\cite{baggiani2026competition}. The short-run event-study result in our companion
paper does not contradict frozen depth as a baseline scenario, but it does not establish
zero LP response; Appendix~\ref{app:m4} relaxes the assumption as a sensitivity layer. Every random draw is
seeded and each exogenous stream (the oracle path, buy and sell noise volumes, arbitrage
participation, and the randomized-ordering coin) is keyed to the episode seed
independently of policy actions, so paired policies on the same seed face identical
primitive shocks; the appendix searcher layer is the one exception (its participation
draws trigger on the policy's own trade sizes) and is disclosed as such. Holding policy, configuration, and software state fixed, an episode
is bit-reproducible from its seed, verified by content hash across all subsequent
refactors.

\subsection{Experimental implementation}
\label{sec:impl}

\paragraph{Observations.} The raw observation $\tilde{o}_t$ exposes pool reserves, mids,
and fees for both pools, the oracle price, remaining inventory and time, per-clip execution
premia, gas, and the previous action; baselines and planners consume it directly and
rebuild the exact quote curves of Eq.~\eqref{eq:cost}. The DQN receives the compressed
vector $o_t = \phi(\tilde{o}_t) \in \mathbb{R}^{16}$: normalized remaining inventory
$R_t/Q$ and time $(H-t)/H$; $\log S_t$; per-pool oracle gaps $(p_{i,t} - S_t)/S_t$ with mid
$p_{i,t} = x_{i,t}/y_{i,t}$; the rival quote gap; the four fees; trailing realized
volatility summarizing $S_{t-w:t}$; estimated execution premia at three clip sizes on the
best route; gas; and the previous action. Reserve depth is not observed directly and
enters $\phi$ only indirectly through the best-route execution premia. The tabular learners
use a discretization $\psi$ over steps-left, remaining inventory, the marginal route gap,
and the best-pool premium ($M = 450$ coarse, $M = 1{,}440$ fine).

\paragraph{Policies.} All ladder policies act in the same eight-action set.
Fastest-feasible liquidation takes a 50\%-of-remaining action each step, routed to the
better-quoted pool (the equal-pace split action is not considered; the rung is a schedule
reference, not an optimized policy). The TWAP policies clip the pace to the largest feasible action fraction,
$\tilde{\alpha}_t = \min\{1/(H - t),\, 0.5\}$, and take the smallest single-pool
fraction in $\{.1, .25, .5\}$ no smaller than $\tilde{\alpha}_t$, routed to the
better-quoted pool (any residual at $H$ is handled by the common completion rule); the fee-aware variant compares
the same-pace candidates (pool $A$, pool $B$, and the split when the pace allows) by
estimated all-in per-unit cost. The lookahead carry is tuned per market mode on
validation seeds and frozen: $\kappa = 16$ in dynamic and constant-fee duopoly,
$\kappa = 8$ in dynamic monopoly. The tuning ran under the standard completion rule
during development; re-running the grid under forced-terminal semantics selects the same
per-mode values, and the per-ordering audit is in Section~\ref{sec:artifacts}. The
deterministic planners use $K \in \{2, 3\}$ with the carry re-tuned per depth (both
depths selected $\kappa = 16$); the stochastic rollout planner selected $K = 3$,
$N = 16$, $\kappa = 16$. The DQN is a
16--64--64--8 MLP with replay buffer of $2 \times 10^5$ transitions, batch 64, Adam
$10^{-3}$, $\varepsilon$ annealed $1.0 \to 0.05$, and $\gamma = 1$; the tabular learners
anneal $\alpha$ and $\varepsilon$ over $2$--$5$ million episodes. Algorithm~\ref{alg:pipeline}
is instantiated with $E = 12{,}000$ training episodes (one fresh seed per episode),
target-network period $\tau = 2{,}000$ steps, evaluation period $E_{\mathrm{eval}} = 500$,
$|\mathcal{S}^{L}_{\mathrm{val}}| = 50$ and $|\mathcal{S}^{B}_{\mathrm{val}}| = 200$, and
$\mathcal{S}_{\mathrm{final}}$ the reserved block of seeds 90{,}000--90{,}999
($n = 1{,}000$), pairwise disjoint from all training, validation, and development blocks.
Because learner design iterated while development test seeds were visible,
$\mathcal{S}_{\mathrm{final}}$ was first evaluated only after the design freeze, with every
choice recorded in a content-hashed run manifest (Appendix~\ref{app:repro}). The main DQN
checkpoint was trained and selected under the standard completion rule
($C^{\mathrm{term}}_H = \rho S_0 R_H$); headline evaluation then switches every policy to
the forced-terminal rule, the uniform semantics change of Section~\ref{sec:objective}.

\paragraph{Failure modes.} Two training details were load-bearing and are reported because
both are audit-relevant. First, \emph{bootstrapped tabular} Q-learning collapsed into
permanent waiting. The mechanism combines two textbook effects under an
all-costs (rewards predominantly negative) objective: zero initialization is effectively
optimistic~\cite[\S 2.6]{sutton2018}, so under-visited actions retain values above their
true negative returns, and the bootstrap of Q-learning~\cite[\S 6.5]{sutton2018}
propagates those optimistic zeros backward through the $\max$ in its target; the waiting
action, whose immediate reward is near zero, keeps $\hat{Q}(\cdot, \mathrm{wait})$ pinned
near zero while every trading action turns negative, so the greedy policy defers
indefinitely. Maximization bias, the upward bias of a $\max$ over noisy estimates
analyzed by van Hasselt~\cite{vanhasselt2010} (see also~\cite[\S 6.7]{sutton2018}), acts
in the same direction. The Monte Carlo target
in Eq.~\eqref{eq:mc} removes the bootstrap and fixes completion (the function-approximation
DQN with a target network did not exhibit the collapse). Second, a premium feature defined
on a fraction of \emph{remaining} inventory confounded market cheapness with position size:
large $R_t$ made every state look expensive, teaching self-trapping passivity, until the
feature was replaced by inventory-independent marginal quotes.

\subsection{Main results}
\label{sec:results}

\begin{figure}[!htbp]
\centering
\begin{subfigure}[b]{0.49\textwidth}
\centering
\includegraphics[width=\textwidth]{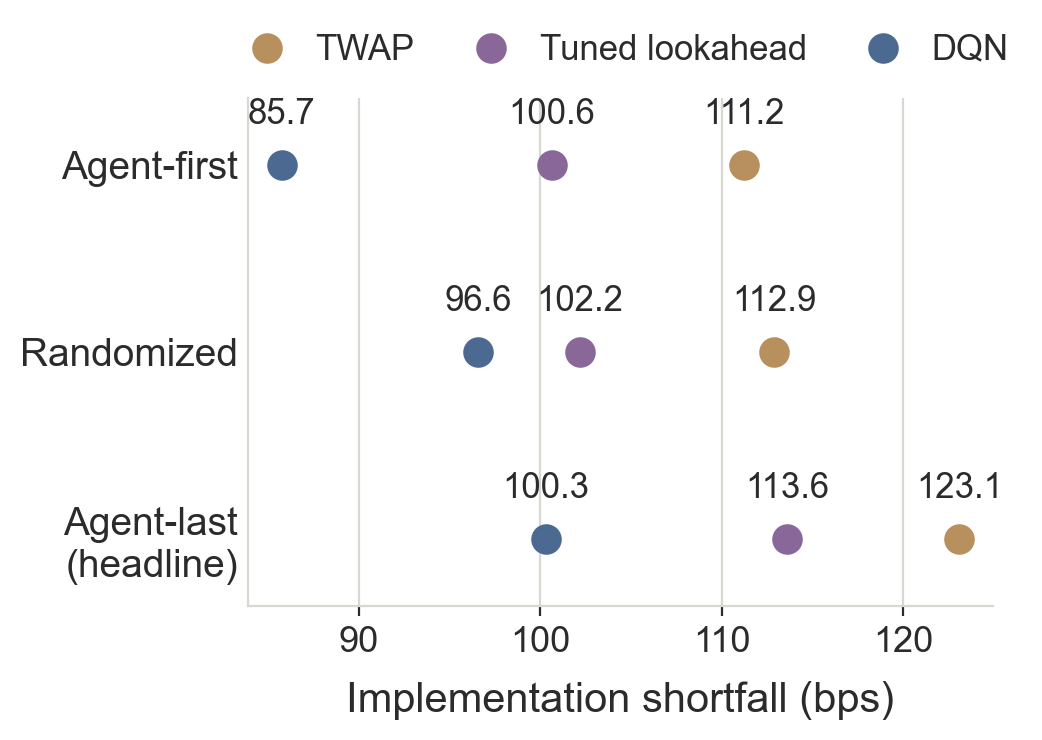}
\caption{Mean shortfall by ordering.}
\label{fig:final-abs}
\end{subfigure}
\hfill
\begin{subfigure}[b]{0.49\textwidth}
\centering
\includegraphics[width=\textwidth]{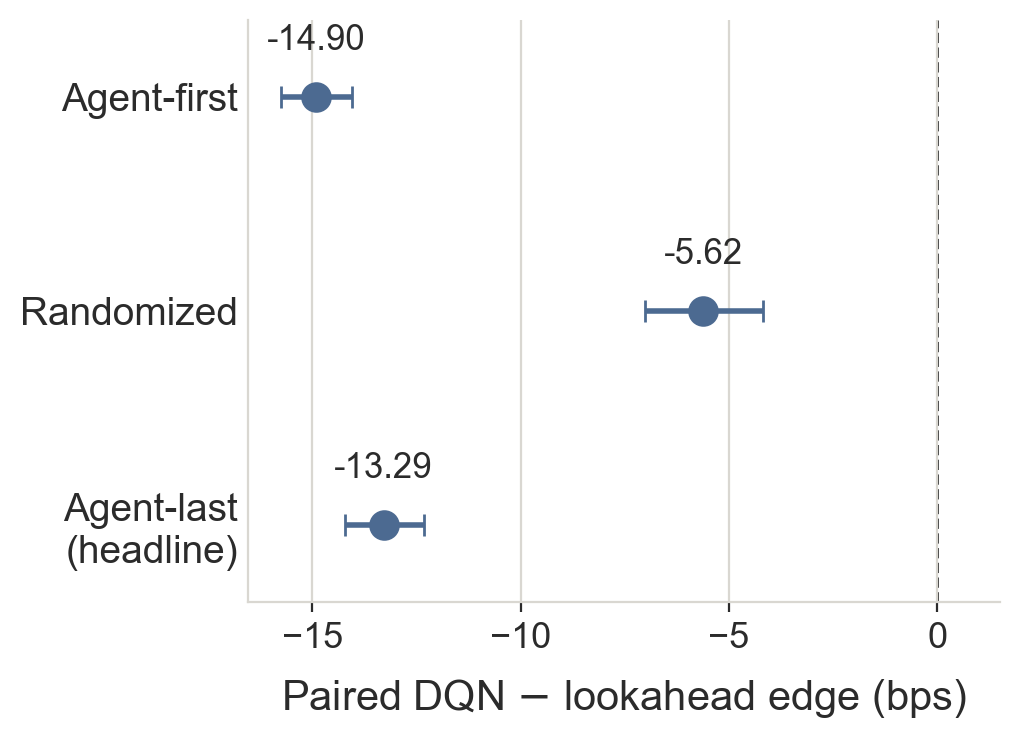}
\caption{Paired DQN $-$ lookahead edge.}
\label{fig:final-edge}
\end{subfigure}
\caption{Final reserved seed block (90{,}000--90{,}999, $n=1{,}000$; first evaluated
after the design freeze), forced-terminal completion, in bps of order notional. Each DQN
is trained under its stated intra-step ordering using the standard completion objective,
then evaluated under forced-terminal completion together with all comparison policies (an
out-of-training-semantics evaluation, Section~\ref{sec:objective}).
(b)~Per-seed paired differences with pointwise 95\% bootstrap CIs ($B = 4{,}000$):
$-14.90$ $[-15.76, -14.05]$ agent-first, $-5.62$ $[-7.03, -4.18]$ randomized (smallest
edge), $-13.29$ $[-14.22, -12.32]$ agent-last (deterministic headline). Negative favors
the DQN; inference is conditional on each frozen checkpoint.}
\label{fig:final}
\end{figure}

Figure~\ref{fig:final} reports the frozen final block: mean shortfall for TWAP, tuned
lookahead, and the DQN under each intra-step ordering, and the paired DQN$-$lookahead
edge with its interval. Under forced-terminal completion every policy completes exactly;
paired intervals are over per-seed differences against lookahead evaluated under the same
ordering.

Three features of the agent-first benchmark ladder (Appendix~\ref{app:ladder}) carry the
interpretation; all paired differences below are against lookahead on the final block,
agent-first ordering, under the same forced-terminal rule. First, planning
does not substitute for learning here: the deterministic two-step planner is \emph{worse}
than tuned one-step control ($+10.8\bps$, CI $[+10.0, +11.6]$), three-step roughly ties
($+1.6$), and the stochastic rollout planner is statistically indistinguishable from one-step
lookahead ($+0.11$, CI $[-0.41, +0.65]$). The results are consistent with exploitable
value in realized fee and quote fluctuations (noise and arbitrage overshoot around the
oracle) that the evaluated shallow planners do not recover; they do not separate this
explanation from limited depth, rollout budget, or the heuristic continuation.
Expectation-based and short sampled rollouts appear to average much of that structure away,
while the tuned carry term already encodes the mean value of deferral. Deeper or
learned-value-guided search was not evaluated and might close part of the gap. Second,
finer discretization removes most of the measured tabular gap: the 450-state learner is
$+8.2\bps$ worse than lookahead, while the 1{,}440-state learner's paired difference is
$-0.42\bps$ with a pointwise $95\%$ CI of $[-1.50, +0.76]$, which contains zero. The DQN is the only evaluated
learner whose improvement excludes zero. Third, the DQN sits within about $7.5\bps$ of the achieved hindsight reference
($78.2\bps$), closing roughly two thirds of the gap between lookahead and the reference.
The reference is a feasible hindsight solution found by coordinate descent, not a
certified optimum, and it consumes realized-shock information no policy can have; it
therefore calibrates measured headroom without certifying optimality of either the
reference or the learner.

\subsection{Policy behavior and mechanism}
\label{sec:behavior}

\begin{figure}[!htbp]
\centering
\begin{subfigure}[b]{0.32\textwidth}
\centering
\includegraphics[width=\textwidth]{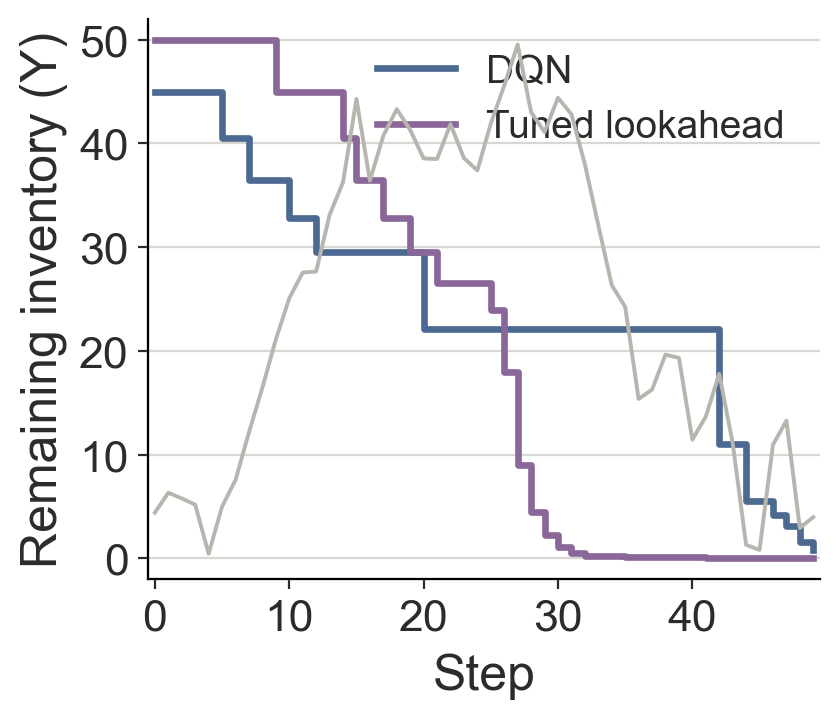}
\caption{DQN wins (seed 30409).}
\label{fig:traj-win}
\end{subfigure}
\hfill
\begin{subfigure}[b]{0.32\textwidth}
\centering
\includegraphics[width=\textwidth]{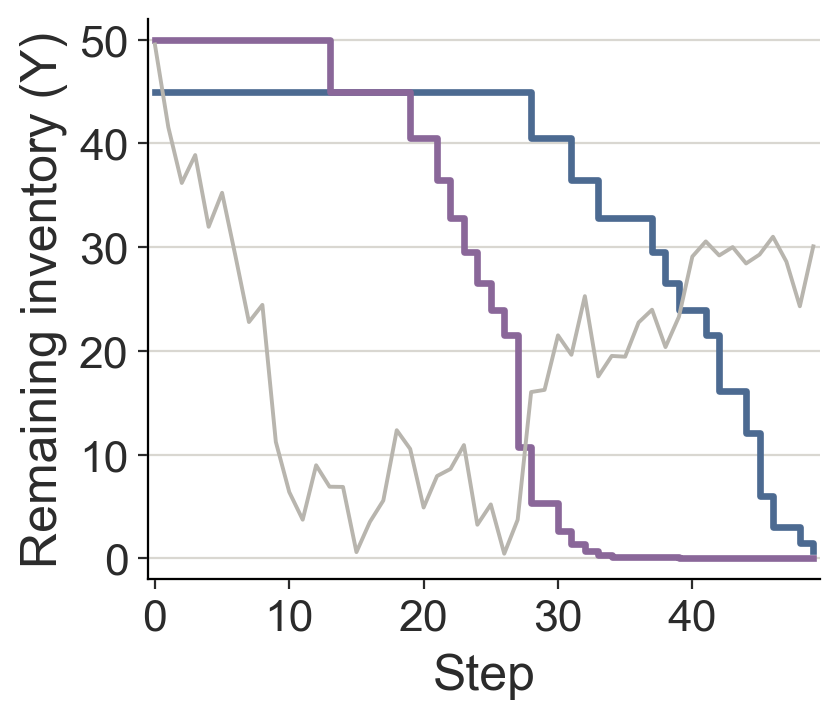}
\caption{DQN loses (seed 30138).}
\label{fig:traj-lose}
\end{subfigure}
\hfill
\begin{subfigure}[b]{0.32\textwidth}
\centering
\includegraphics[width=\textwidth]{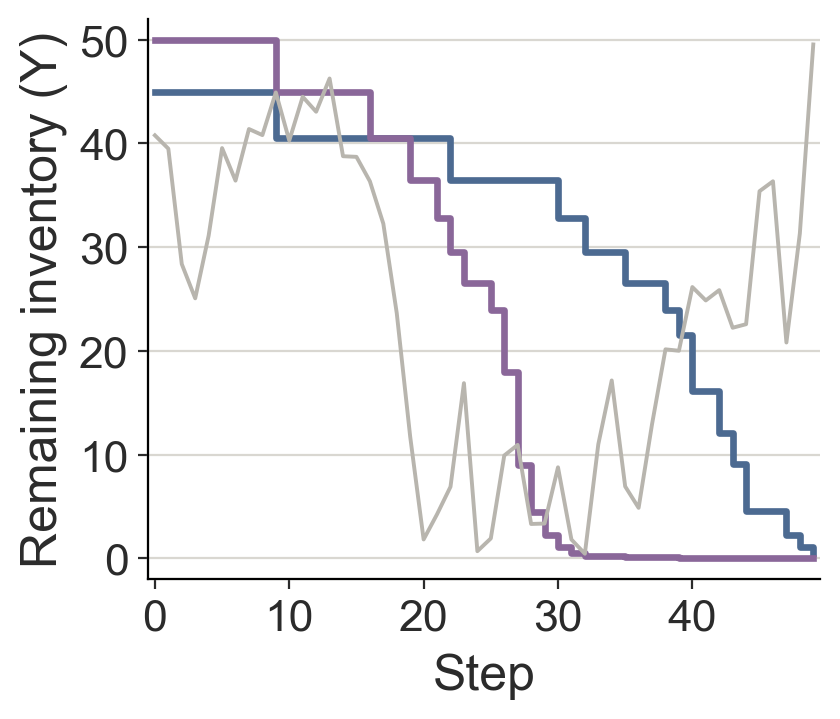}
\caption{Tie (seed 30266).}
\label{fig:traj-tie}
\end{subfigure}
\caption{Representative episodes: remaining inventory for the DQN and tuned lookahead,
with the oracle price path in gray (right axis, unscaled). The DQN paces small clips and
defers through price run-ups; lookahead concentrates execution mid-episode. The losing seed
shows the failure mode: the DQN waits for dips that never come.}
\label{fig:traj}
\end{figure}

\begin{figure}[!htbp]
\centering
\begin{subfigure}[b]{0.49\textwidth}
\centering
\includegraphics[width=\textwidth]{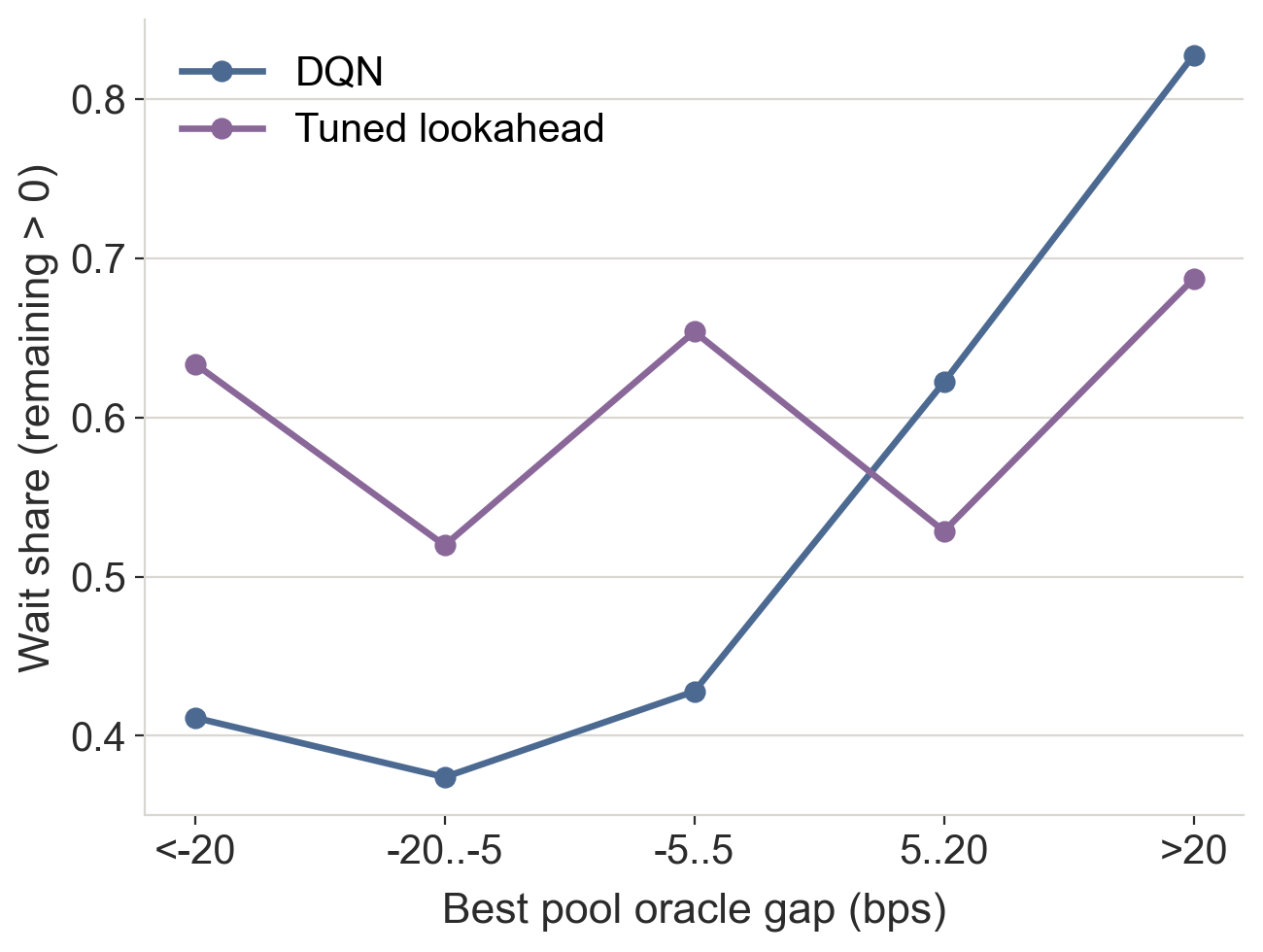}
\caption{Waiting vs.\ oracle gap.}
\label{fig:behavior-wait}
\end{subfigure}
\hfill
\begin{subfigure}[b]{0.49\textwidth}
\centering
\includegraphics[width=\textwidth]{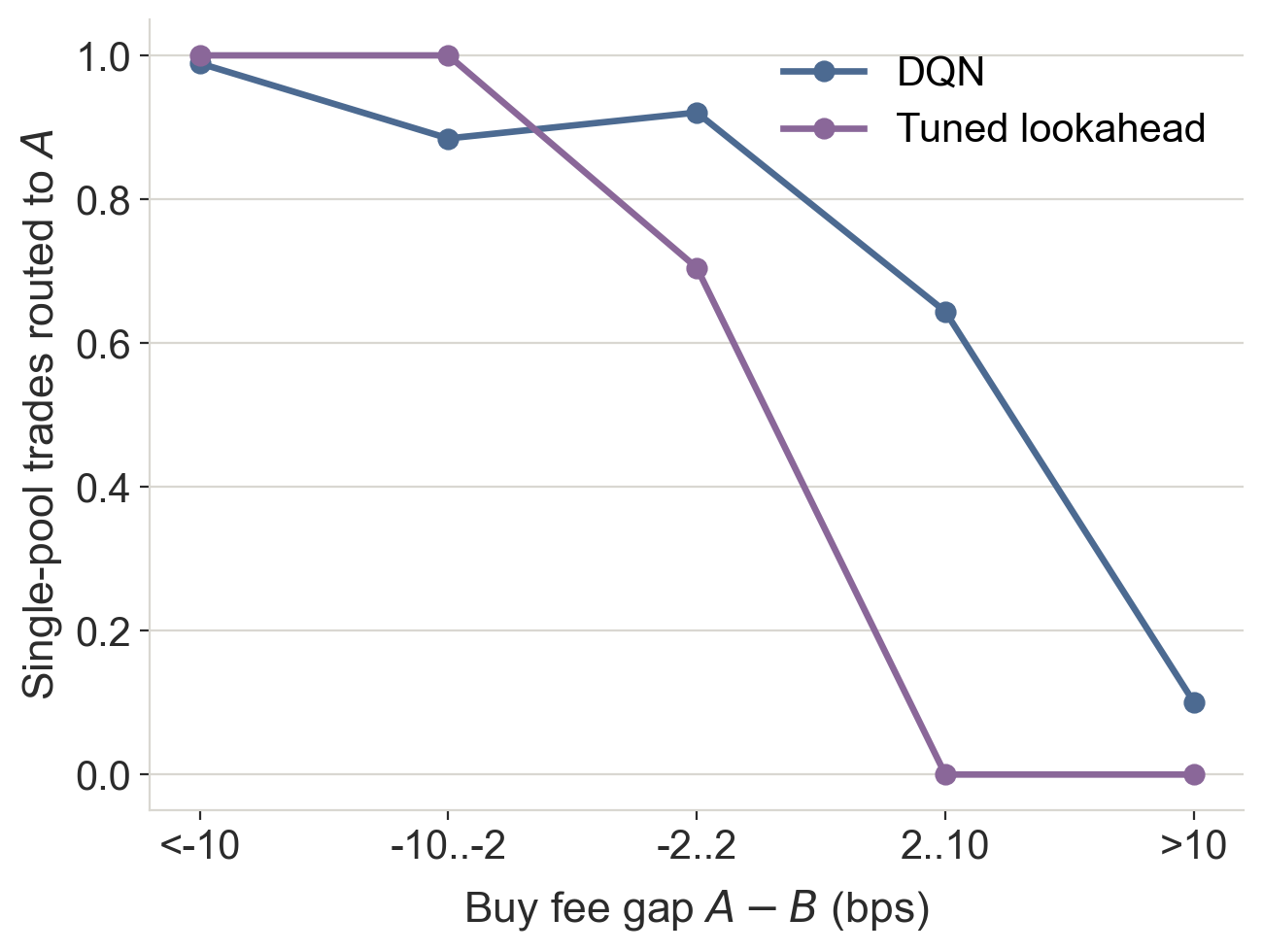}
\caption{Routing vs.\ fee gap.}
\label{fig:behavior-routing}
\end{subfigure}
\caption{State-conditional behavior on development test seeds, DQN vs.\ tuned lookahead.
(a)~Wait share against the best pool's oracle gap (negative gap = pool cheap relative to
the oracle); the DQN's wait share is monotone in the gap. (b)~Share of single-pool trades
routed to pool $A$ against the buy-fee gap $A-B$ (negative = $A$ cheaper); the DQN's
residual preference for pool $A$ at small fee gaps is the costless symmetry break verified
by the label-swap check.}
\label{fig:behavior}
\end{figure}

On development test seeds (standard completion rule, agent-first ordering), the
penalized-score decomposition (Section~\ref{sec:objective}) locates the DQN's advantage
over lookahead in executed-price quality:
slippage-ex-fee $-11.1\bps$, fees $-4.3$, gas $-2.1$, drift $-1.2$, against $+4.9$ of terminal
penalty under the standard rule. Mechanically the policy trades roughly sixteen 10\%-of-remaining
clips, defers on oracle run-ups and buys dips, and times trades into low-fee states: realized pool-fee cash contributes $33.5\bps$ of
arrival-notional shortfall, while the start-of-step quoted buy fee averages $45\bps$
across pool-time states; these are different summaries, and their difference is not a
direct estimate of fee savings. Routing follows the fee gap
almost logistically for learner and heuristic alike (Figure~\ref{fig:behavior}). The failure
mode is symmetric: on seeds where pools cheapen early and never again, deferral loses to
lookahead's earlier liquidation (Figure~\ref{fig:traj}).

\subsection{Robustness}
\label{sec:artifacts}

\paragraph{Completion.} Under the standard rule the unconstrained DQN completes $0.9755$ on
average. This is not the source of the edge: unfilled inventory is charged the terminal
penalty $10^4 \rho\, R_H / Q = 200\,(R_H/Q)\bps$ of order-level shortfall, a rate two to
three times the marginal cost of completing, and
switching every policy to forced execution \emph{lowers} the DQN's reported number
($85.9 \to 84.9\bps$ on development seeds; a penalized score and a realized shortfall
under different terminal semantics, Section~\ref{sec:objective}) because the forced leg
costs less than the penalty it replaces; since the penalty term is absent under the
forced rule, the surviving edge cannot be penalty-avoidance behavior. All headline numbers use
the forced rule for every policy. A retrain with a quadrupled training-time penalty reached
$0.990$ completion and is dominated; it is reported as robustness only.

\begin{figure}[!htbp]
\centering
\begin{subfigure}[b]{0.49\textwidth}
\centering
\includegraphics[width=\textwidth]{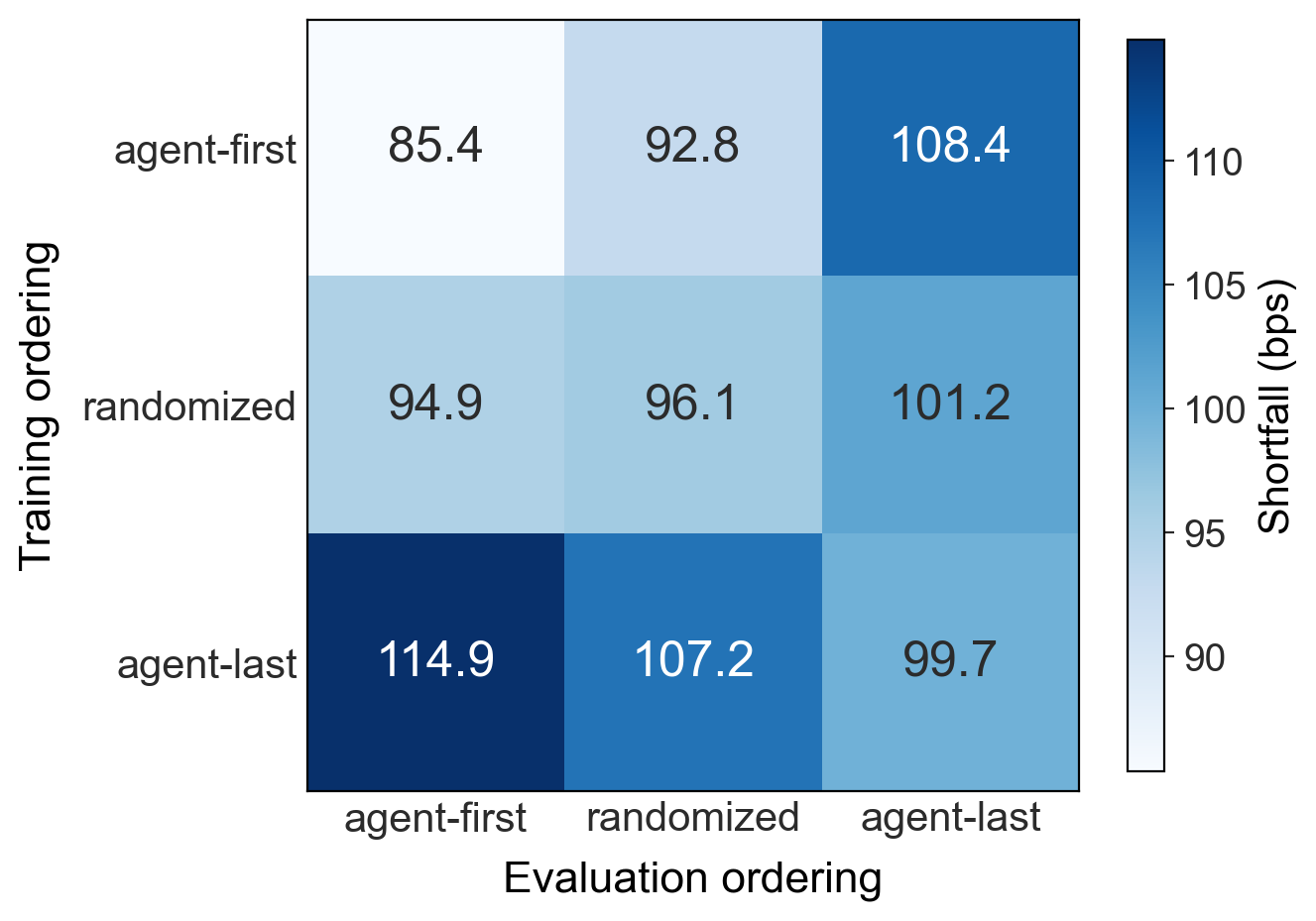}
\caption{DQN shortfall (bps).}
\label{fig:priority-abs}
\end{subfigure}
\hfill
\begin{subfigure}[b]{0.49\textwidth}
\centering
\includegraphics[width=\textwidth]{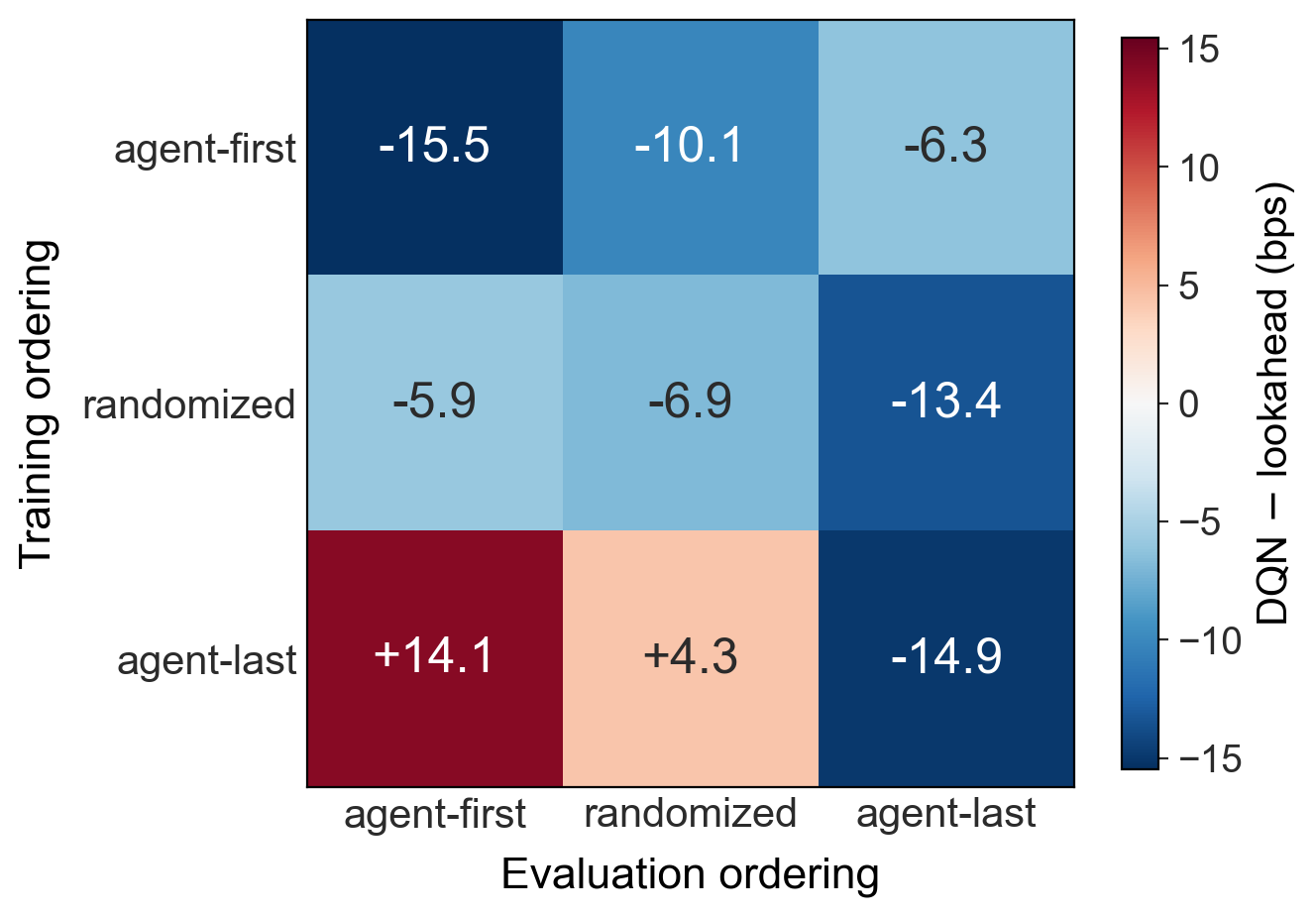}
\caption{Paired difference vs.\ lookahead.}
\label{fig:priority-edge}
\end{subfigure}
\caption{Priority retraining on 300 development test seeds, forced-terminal completion.
(a)~DQN shortfall by training and evaluation ordering. (b)~Paired difference against
lookahead evaluated under the same ordering (negative favors the DQN). Diagonal cells
(trained and tested under the same ordering) beat lookahead everywhere; policies specialize
to their training ordering.}
\label{fig:priority}
\end{figure}

\paragraph{Priority ordering.} Intra-step priority initially looked like a vulnerability:
the agent-first edge of $-13.6\bps$ shrank to $-4.1$ when the frozen policy was evaluated
agent-last. Retraining under each ordering resolves it (Figure~\ref{fig:priority}):
trained-and-tested agent-last, the DQN beats same-ordering lookahead by $-13.3\bps$ on the
final block; randomized ordering yields $-5.6$. The recovery after same-order retraining
is consistent with policy--environment mismatch being an important component of the
transfer loss, though it does not rule out ordering-specific execution rents: the
transfer-versus-re-solving distinction of Section~\ref{sec:transfer} in action. Randomized ordering is the learner's hardest regime,
which is informative in itself: when the agent cannot know whether its trade lands before
or after this step's noise and arbitrage, the map from observed quote to execution price
acquires irreducible noise, and fine timing is worth less. Lookahead is not retrained
across orderings, but its realized cost still changes with execution order
($100.6$ / $102.2$ / $113.6\bps$ in Figure~\ref{fig:final}, all at the frozen
$\kappa = 16$); the DQN's additional value
depends on learning the cost-to-go under the specific ordering, which is why same-order
retraining matters, and the paper reports both the deterministic headline and the smallest
(randomized) edge for that reason. The carry was tuned per market mode under the default
agent-first ordering and frozen at $\kappa = 16$ for dynamic duopoly before any
cross-ordering or final-block run; every reported row uses that pre-specified comparator,
and no comparator choice was revisited after the final block was opened. A post-hoc
per-ordering audit of the validation grid (forced-terminal semantics, validation seeds
only) selects the frozen $\kappa = 16$ under agent-first and agent-last, and prefers
$\kappa = 8$ under randomized ordering; as a sensitivity, evaluating the randomized cell
against that re-tuned $\kappa = 8$ lookahead \emph{widens} the paired edge to $-7.48$
(CI $[-8.61, -6.30]$), so the randomized conclusion does not depend on which comparator
is used.

\paragraph{Fee-mode attribution.} Retraining the DQN per market mode separates dynamic-fee
structure from generic closed-loop timing. Because the carry is tuned per mode
(Section~\ref{sec:impl}), each in-mode comparison uses that mode's validation-selected
lookahead: $\kappa = 16$ in constant and dynamic duopoly, $\kappa = 8$ in monopoly. In
constant-fee duopoly the in-mode DQN's estimated paired difference against in-mode
lookahead is $+0.39\bps$ with a pointwise $95\%$ CI of $[-2.60, +3.58]$, which includes
zero ($116.1$ vs. $115.8\bps$ in means); under the dynamic rule the edge is $-15.5$ in
duopoly and $-23.5$ (CI $[-25.2, -21.8]$) in monopoly ($-27.4$ against a $\kappa = 16$
variant). The mechanism is visible in behavior
(Section~\ref{sec:behavior}): the own-pool and oracle terms of Eq.~\eqref{eq:fee} both
lower the buy fee, all else equal, when a pool sits below the oracle, so cheap-fee and
cheap-quote windows tend to coincide and are worth timing. In the matched model ablation the estimated DQN$-$lookahead edge is smaller in dynamic
duopoly than in dynamic monopoly. This is consistent with competition reducing the amount
of exploitable fee-state variation, although the experiment does not identify that
mechanism separately, and a policy-versus-benchmark gap is not a welfare measure in the
sense of~\cite{baggiani2026competition}. One nuance is disclosed rather than
resolved: the duopoly-trained policy transfers a $-8.9\bps$ edge \emph{into} constant-fee
mode, while the constant-trained policy finds nothing anywhere; the asymmetry is
consistent with dynamic-fee training acting as a useful curriculum, though this mechanism
is not separately identified. Accordingly the claim is
that the edge
is concentrated in, and learned from, dynamic-fee environments, not that dynamic fees
cause it.

\paragraph{Parameter perturbations.} The remaining battery, run on the frozen policy: gas
at $0/2/10\,X$ preserves the edge and widens it at high gas; arbitrage speed at
$0.2/0.5/1.0$ preserves it at all speeds; doubled noise intensity and doubled volume
dispersion preserve it; perturbing each fee coefficient by $\times 0.5$ and $\times 2$
never flips a ranking (the oracle term is the load-bearing one); and the fee rule exhibits
the expected two-regime tilt. Swapping pool labels in observation and action reproduces the
shortfall bit-for-bit with the route share mirrored ($0.78 \to 0.22$), so the learner's
pool concentration is an arbitrary symmetry break, not hidden asymmetry. The observation is
guarded by a schema whitelist test, and the shortfall decomposition (drift, slippage-ex-fee,
fee, gas, terminal) satisfies an exact accounting identity enforced by a regression test.

\FloatBarrier
\section{Scope and Conclusion}
\label{sec:scope}

The result is model-conditioned. It supports: in this environment, model-free execution control
beats tuned one-step routing under strict completion and the agent-last deterministic ordering; the
edge is concentrated in and learned from dynamic-fee environments; shallow deterministic and
stochastic planning, as evaluated here, does not reach it. It does not support: profitability
of any deployed
strategy; identification of historical trader behavior (the event-study boundary is untouched;
historical data enters only as calibration scale); Nash equilibrium of the fee rule (closure
only); or optimality of dynamic fees for liquidity providers. Within the sensitivity models of
Appendix~\ref{app:m4}, the edge survives LP withdrawal and a static sandwich layer, but those
appendices are explicit models, not statements about general MEV or real LP behavior.

All performance claims are conditional on the present simulator, feature map, completion
semantics, and fee closure: a researcher-specified 16-feature compression, an eight-action
discretization, a risk-neutral expected episode-score objective (with forced-terminal
shortfall as the headline metric), and a clipped linear fee
rule. We do not identify which features are necessary, evaluate continuous or
risk-sensitive control, or test nonlinear fee mechanisms. Forced-terminal evaluation
represents a must-complete parent order; optional cancellation defines a different control
problem requiring an explicit non-execution utility.

The missing sequential signal was not hidden in the historical tape; it had to be created by
closing the market loop. Once closed --- with equilibrium-inspired dynamic fees as the market's
reflex rather than as a solved object --- the environment contains measured
within-model execution value: schedule baselines leave $\sim 11\bps$ to tuned one-step
control, tuned control leaves $5$--$15\bps$ to a small model-free learner depending on
intra-step priority, and, in the agent-first ladder, the learner sits within
$\sim 7.5\bps$ of the achieved hindsight reference. The method of Section~\ref{sec:method}
(an action-responsive environment, common completion semantics, a tuned policy ladder, a
frozen reserved block, and the transfer-versus-re-solving distinction) is what turned an
easy-to-overstate ``RL beats X'' number into a claim with explicit conditions, and it can
be reused in other simulators with simulator-specific semantics and artifact checks. Further work should separate feature dependence from learning and evaluate nonlinear fee
closures, continuous sizing, and risk-sensitive execution, alongside richer liquidity and
searcher dynamics, each gated behind the same discipline.

\FloatBarrier
\appendix

\section{Agent-first benchmark ladder}
\label{app:ladder}

\begin{figure}[!htbp]
\centering
\includegraphics[width=0.86\textwidth]{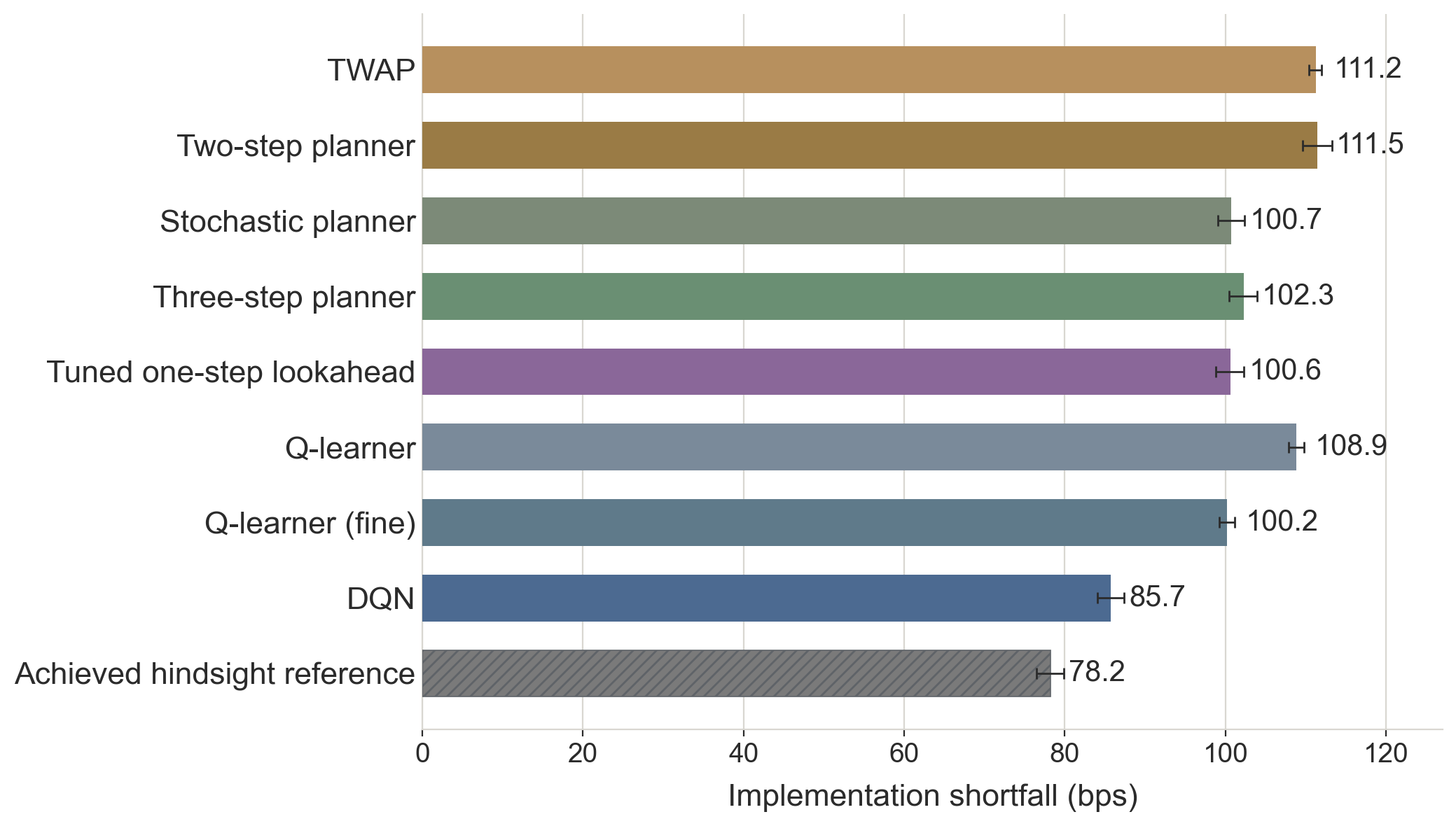}
\caption{Agent-first benchmark ladder on the final reserved seed block (90{,}000--90{,}999,
$n = 1{,}000$; DynamicDuopoly, forced-terminal completion, agent-first ordering; 95\%
bootstrap CIs; completion is $1.0$ for every policy). Schedule baselines and the two-step
planner sit at the top, tuned one-step lookahead at $100.6\bps$, the coarse tabular
learner above lookahead and the finer learner's point estimate slightly below it with a
paired interval spanning zero, the DQN at
$85.7\bps$, and the achieved hindsight reference at $78.2\bps$ (hatched: not a
deployable policy).
Among the evaluated valid policies, only the DQN has a paired $95\%$ interval lying
entirely below zero relative to lookahead; it closes roughly two thirds of the gap
between lookahead and the reference.}
\label{fig:ladder}
\end{figure}

Figure~\ref{fig:ladder} evaluates the full policy ladder on the final reserved seed block
under the agent-first ordering used throughout development; the deterministic headline in
Figure~\ref{fig:final} is the agent-last cell. The ladder is reported in this single
ordering because the planners, tabular learners, and hindsight reference were developed and
tuned agent-first; the DQN row is the agent-first checkpoint. Cross-ordering comparisons
belong to Section~\ref{sec:artifacts} and Figure~\ref{fig:final}.

\FloatBarrier
\section{LP-adaptation and searcher sensitivity}
\label{app:m4}

\begin{figure}[!htbp]
\centering
\includegraphics[width=0.8\textwidth]{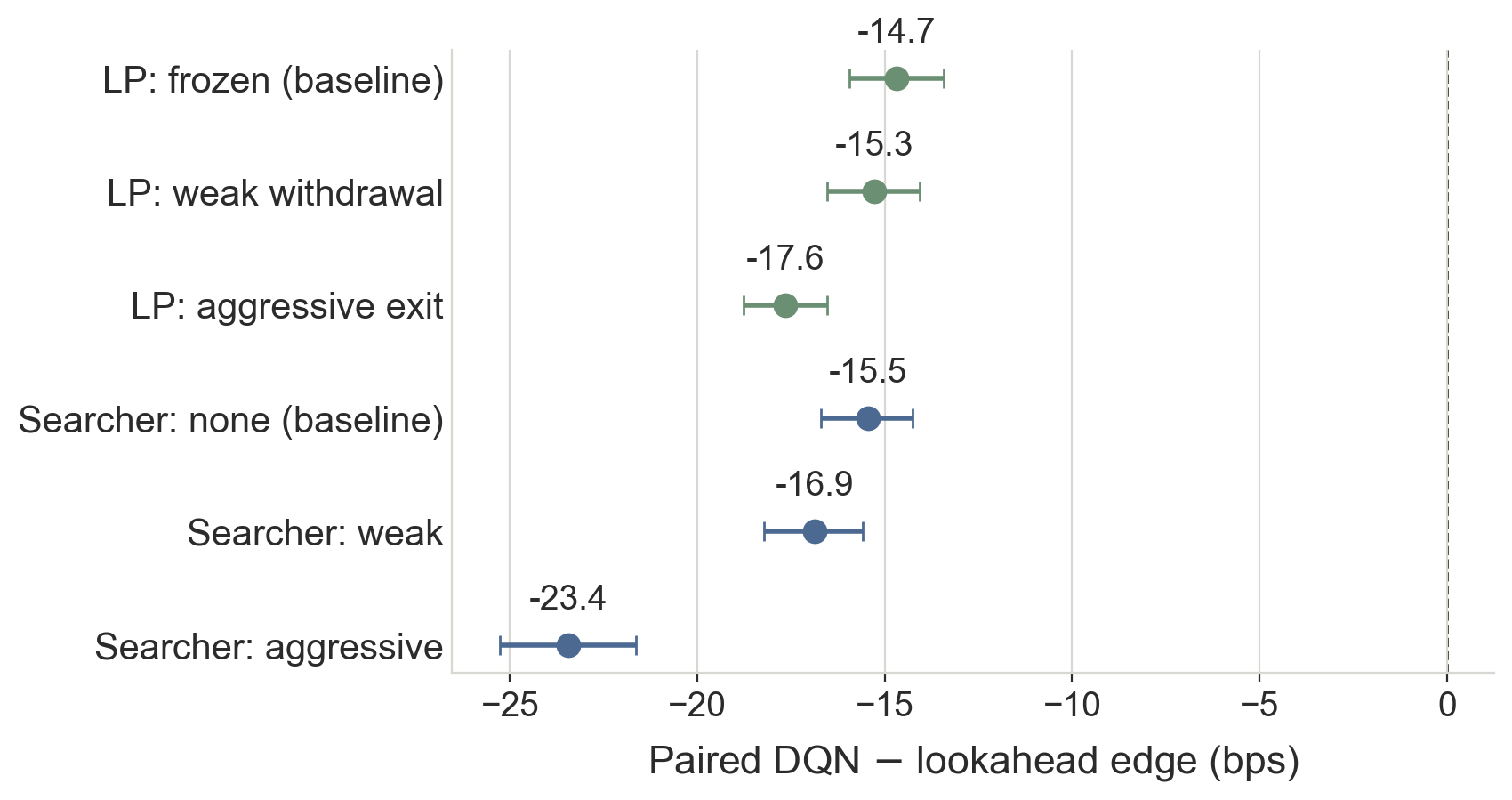}
\caption{Sensitivity layers: paired DQN $-$ lookahead edge (points) with pointwise 95\%
bootstrap CIs (whiskers), 500-seed reserved blocks, frozen agent-first checkpoint, no
retraining. Every interval lies below zero. Depth troughs under LP adaptation are $0.97$
(weak) and $0.58$ (aggressive); under the aggressive searcher, absolute costs rise by
roughly $70\bps$ for every policy.}
\label{fig:sensitivity}
\end{figure}

Both layers are additive and default-off; a unit test pins the defaults to the exact frozen
semantics. \emph{LP adaptation} scales both reserves of a pool by a common factor
(price-preserving) in response to toxic flow, proxied per pool by arbitrage execution in
a trailing 10-step window (the flow component behind
loss-versus-rebalancing~\cite{milionis2022lvr}): when the window holds at least $3$ (weak)
or $2$ (aggressive) arbitrage hits, the depth factor multiplies by $0.997$ ($-0.3\%$,
weak) or $0.98$ ($-2\%$, aggressive); otherwise it recovers by $\times 1.001$ (weak) or
$\times 1.0005$ (aggressive), always clamped to $[\text{floor}, 1]$ with floors $0.90$
and $0.50$. The regimes are reduced-form; optimal LP behavior has its own control
literature~\cite{cartea2024predictable} and is not modeled. \emph{Searcher} sandwiches
agent trades above a size threshold, the canonical front-run/back-run pattern
of~\cite{daian2020flashboys}: when the agent's per-pool quantity $q_i$ reaches the
threshold, a single independent Bernoulli draw (probability $p$) for that eligible agent
leg determines whether the paired front-run and back-run are both applied; if so, the
searcher buys $\phi\, q_i$ on the same pool immediately before the agent's leg and sells
the same quantity back immediately after it, paying pool fees on both legs, with no gas
charge and no profitability test (an unconditional stress, not an optimizing searcher). Regimes:
weak (threshold $5\,Y$, fraction $\phi = 0.25$, $p = 0.5$) and aggressive (threshold
$2\,Y$, $\phi = 1.0$, $p = 1.0$); the layer is a
deliberate simplification of the Stackelberg bot game of~\cite{bayraktar2024dex}, retained
only as stress. On reserved
500-seed blocks with forced-terminal completion (completion $1.0$ in every cell), the
paired DQN$-$lookahead edge survives every regime
(Figure~\ref{fig:sensitivity}).

Thinner books penalize larger clips more, so the small-clip policy degrades gracefully under LP
exit. Under the aggressive searcher, all of the damage lands in slippage-ex-fee (the sandwich
channel), and the baseline ranking reshuffles: TWAP's smaller schedule is less exposed to the
threshold rule than lookahead (9.0 vs. 11.8 sandwiches per episode) and overtakes it, although
late and forced-terminal clips still trigger sandwiches; the learner stays ahead throughout.
A real searcher would adapt its threshold, and no searcher-aware retraining was done; the
regimes are stress models, not MEV-market estimates.

\FloatBarrier
\section{Parameter provenance and reproducibility}
\label{app:repro}

\begin{table}[!htbp]
\centering
\caption{Parameter provenance. Classes: empirically anchored scale (order of magnitude
chosen to match common on-chain magnitudes; no formal estimation); literature-inspired
(functional form from cited work, value chosen); researcher-chosen and frozen before the
design freeze; validation-selected (Algorithm~\ref{alg:pipeline}, step 3); robustness
axis (perturbed in Section~\ref{sec:artifacts} or Appendix~\ref{app:m4}).}
\label{tab:provenance}
\footnotesize
\begin{tabular}{>{\raggedright\arraybackslash}p{0.34\textwidth}>{\raggedright\arraybackslash}p{0.2\textwidth}>{\raggedright\arraybackslash}p{0.36\textwidth}}
\toprule
Parameter & Value & Provenance \\
\midrule
Base fee $\bar{f}$ & $30\bps$ & empirically anchored scale (common fee tier) \\
Oracle volatility $\sigma$ (annualized) & $0.5$ & empirically anchored scale \\
Order / depth; horizon & $Q{=}50\,Y$ (5\% of one pool); $H{=}50$ & empirically anchored
scale; frozen \\
Agent gas $g$ & $2\,X$ per pool & empirically anchored scale; robustness axis \\
Pool reserves & $10^6\,X$, $10^3\,Y$ & researcher-chosen (normalization) \\
Initial oracle price $S_0$ & $1{,}000\,X/Y$ & researcher-chosen; pools start
oracle-aligned ($z_{i,0} = m_{i,0} = 0$) \\
Fee coefficients $(a_{\mathrm{own}}, a_{\mathrm{riv}}, a_{\mathrm{orc}})$ &
$(0.01, -0.005, 0.5)$ & form literature-inspired~\cite{baggiani2026competition}; value
researcher-chosen; robustness axis \\
Fee clip $[f_{\min}, f_{\max}]$ & $[1, 500]\bps$ & researcher-chosen design parameter \\
Terminal penalty rate $\rho$ & $0.02$ & researcher-chosen; audited (completion) \\
CEX fee; arb.\ gas; arb.\ speed & $10\bps$; $5\,X$; $1.0$ & researcher-chosen; speed is a
robustness axis \\
Noise intensities / dispersion / routing & App.~\ref{app:repro} & researcher-chosen;
robustness axis \\
Volatility window $w$ & $20$ steps & researcher-chosen \\
Lookahead $\kappa$; planner $(K, N, \kappa)$ & mode-specific $8$ or $16$; $(3, 16, 16)$ &
validation-selected per mode; frozen \\
DQN architecture and optimizer & Sec.~\ref{sec:impl} & researcher-chosen and frozen;
checkpoint validation-selected \\
Observation / action design & $d = 16$; eight discrete actions & researcher-chosen and
frozen; feature necessity not identified \\
Objective & expected episode score; $\gamma = 1$ & risk-neutral; standard rule in
training, forced-terminal IS in headline evaluation \\
LP-adaptation / searcher regimes & App.~\ref{app:m4} & robustness-only \\
\bottomrule
\end{tabular}
\end{table} 
\paragraph{Stochastic primitives.} For completeness, the market operators in closed form.
Oracle (zero-drift GBM at one-minute resolution, $\Delta = 1/525{,}600$ years, started at
$S_0 = 1{,}000$):
\[
S_{t+1} = S_t \exp\!\big(-\tfrac{1}{2}\sigma^2 \Delta + \sigma \sqrt{\Delta}\,
\varepsilon_t\big), \qquad \varepsilon_t \sim \mathcal{N}(0, 1)\ \text{i.i.d.}
\]
Noise volumes (uniform-in-exponent multiplicative noise):
\[
V^{\pm}_t = \lambda^{\pm} \exp(\eta\, U^{\pm}_t), \qquad
U^{\pm}_t \sim \mathrm{Unif}[-1, 1],
\]
with median volume scales $\lambda^{\pm} = 2.0\,Y$ and dispersion $\eta = 0.5$ (the mean
volume is $\lambda\, \sinh(\eta)/\eta \approx 2.08\,Y$; the deterministic planner's
forward model uses this exact mean). Routing of each
volume between the two pools is logistic in the effective-cost gap of a $1\,Y$ probe,
\[
w_A = \big(1 + e^{-\beta\,(c_B - c_A)}\big)^{-1},
\]
with $c_i$ in bps of the oracle price and sensitivity $\beta = 0.05$ per bps; each volume
is split deterministically, $V_A = w_A V$ and $V_B = (1 - w_A) V$ (a fractional split,
not a categorical draw). Randomized
priority draws $B_t \sim \mathrm{Bernoulli}(1/2)$ per step; $B_t = 1$ places the agent
leg first, $B_t = 0$ places it after the arbitrage response. The arbitrageur, with
constant-product invariant $k = x_i y_i$, mid $p_i = x_i / y_i$, CEX fee
$f_{\mathrm{cex}}$, and per-transaction gas $g_{\mathrm{arb}}$: if
$p_i < p^{\mathrm{buy}} := S_t (1 - f_{\mathrm{cex}})(1 - f^{\mathrm{buy}}_i)$ it buys
$\Delta y = y_i - \sqrt{k / p^{\mathrm{buy}}}$ from the pool; if
$p_i > p^{\mathrm{sell}} := S_t (1 + f_{\mathrm{cex}}) / (1 - f^{\mathrm{sell}}_i)$ it
sells $\Delta y = \sqrt{k / p^{\mathrm{sell}}} - y_i$ to the pool. The buy leg pays the
cash cost $c_i(\Delta y)$ of Eq.~\eqref{eq:cost}, for profit
$\Pi^{\mathrm{buy}} = S_t (1 - f_{\mathrm{cex}})\, \Delta y - c_i(\Delta y)
- g_{\mathrm{arb}}$. The sell leg mirrors the buy convention with the fee withheld on the
numeraire side: the full $\Delta y$ enters the reserves, the pool releases gross proceeds
$\Delta x = x_i \Delta y / (y_i + \Delta y)$ (so $x_i \leftarrow x_i - \Delta x$,
$y_i \leftarrow y_i + \Delta y$ preserves $k$ exactly and lands the mid on
$p^{\mathrm{sell}}$), and the trader receives $(1 - f^{\mathrm{sell}}_i)\, \Delta x$, for
profit $\Pi^{\mathrm{sell}} = (1 - f^{\mathrm{sell}}_i)\, \Delta x
- S_t (1 + f_{\mathrm{cex}})\, \Delta y - g_{\mathrm{arb}}$. The trade executes only if
the realized profit is positive, and, when the participation probability is below one,
only in steps where an independent coin permits.

Environment constants not shown in Table~\ref{tab:provenance}: noise buy and sell
median volume scales $2.0\,Y$ per step with multiplicative volume dispersion $0.5$
(uniform-in-exponent); softmax routing sensitivity $0.05$ per bps of effective-cost gap,
probed at a $1\,Y$ clip. DQN training constants: replay buffer $2 \times 10^5$, batch 64,
Adam $10^{-3}$, Huber loss, $\varepsilon: 1.0 \to 0.05$, $E = 12{,}000$, $\tau = 2{,}000$,
$E_{\mathrm{eval}} = 500$, CPU-only with fixed framework seed.

The simulator is bit-deterministic per seed (content-hash verified across all refactors since
the environment freeze). A run manifest records interpreter, library, and compiler versions,
OS and CPU, the git commit, the full command list, the seed protocol, frozen design choices,
and SHA-256 hashes of every learner checkpoint and result file. No reinforcement-learning
framework is used: the simulator, the tabular learners, and all planners are implemented
directly in Rust; the DQN is implemented directly in PyTorch (used only as a tensor and
autograd library), and the environment bridge exposes a Gymnasium-style
\texttt{reset}/\texttt{step} interface over JSON lines without depending on Gymnasium. Two
regression tests guard the evaluation semantics: an observation schema whitelist (any new field fails until reviewed for
future-information leakage) and the exact shortfall decomposition identity across seeds,
policies, and completion rules. Training is CPU-only, single-threaded, and seeded. Checkpoints and result files are not
distributed; the repository pipeline regenerates them from the recorded seed protocol, and
the manifest's content hashes allow a regenerated run to be checked against the reported
one (exact retraining reproducibility holds on the recorded toolchain; evaluation from a
regenerated checkpoint follows the same frozen protocol). Code, runners, and the artifact
verifier are available at \url{https://github.com/egpivo/amm-lab}.

\end{document}